\documentclass{article}

\usepackage[preprint]{neurips_2026}

\usepackage[utf8]{inputenc}
\usepackage[T1]{fontenc}
\usepackage{microtype}
\usepackage{amsmath,amssymb,amsfonts}
\usepackage{booktabs}
\usepackage{array}
\usepackage{tabularx}
\usepackage{graphicx}
\usepackage{xcolor}
\usepackage{caption}
\usepackage{flafter}
\usepackage{placeins}
\usepackage{hyperref}
\usepackage{url}

\hypersetup{colorlinks=true,linkcolor=black,citecolor=black,urlcolor=black,pdfauthor={Songhe Wang, David Miller, Charles A. Kamhoua},pdftitle={Deformable Contact-Aware 3D Object Placement}}
\renewcommand{\arraystretch}{1.1}
\newcommand{\method}{DCAP}

\newcolumntype{Y}{>{\raggedright\arraybackslash}X}
\newcolumntype{P}[1]{>{\raggedright\arraybackslash}p{#1}}
\newcommand{\figurebox}[2]{%
\begin{figure}[t]
\centering
\fbox{\begin{minipage}[c][#1][c]{0.92\linewidth}
\centering
\textbf{#2}
\end{minipage}}
}
\definecolor{dcapgreen}{RGB}{42,112,31}
\definecolor{failred}{RGB}{190,18,18}
\definecolor{panelgray}{RGB}{205,210,216}
\definecolor{softgreen}{RGB}{245,251,244}
\definecolor{softred}{RGB}{255,246,245}

\providecommand{\na}{n.a.}
\providecommand{\metricup}{$\uparrow$}
\providecommand{\metricdown}{$\downarrow$}

\title{Deformable Contact-Aware 3D Object Placement}
\author{Songhe Wang\qquad David Miller\qquad Charles A.~Kamhoua}
\date{}

\begin{document}
\maketitle

\begin{center}
\vspace{-0.6em}
\includegraphics[width=0.92\linewidth]{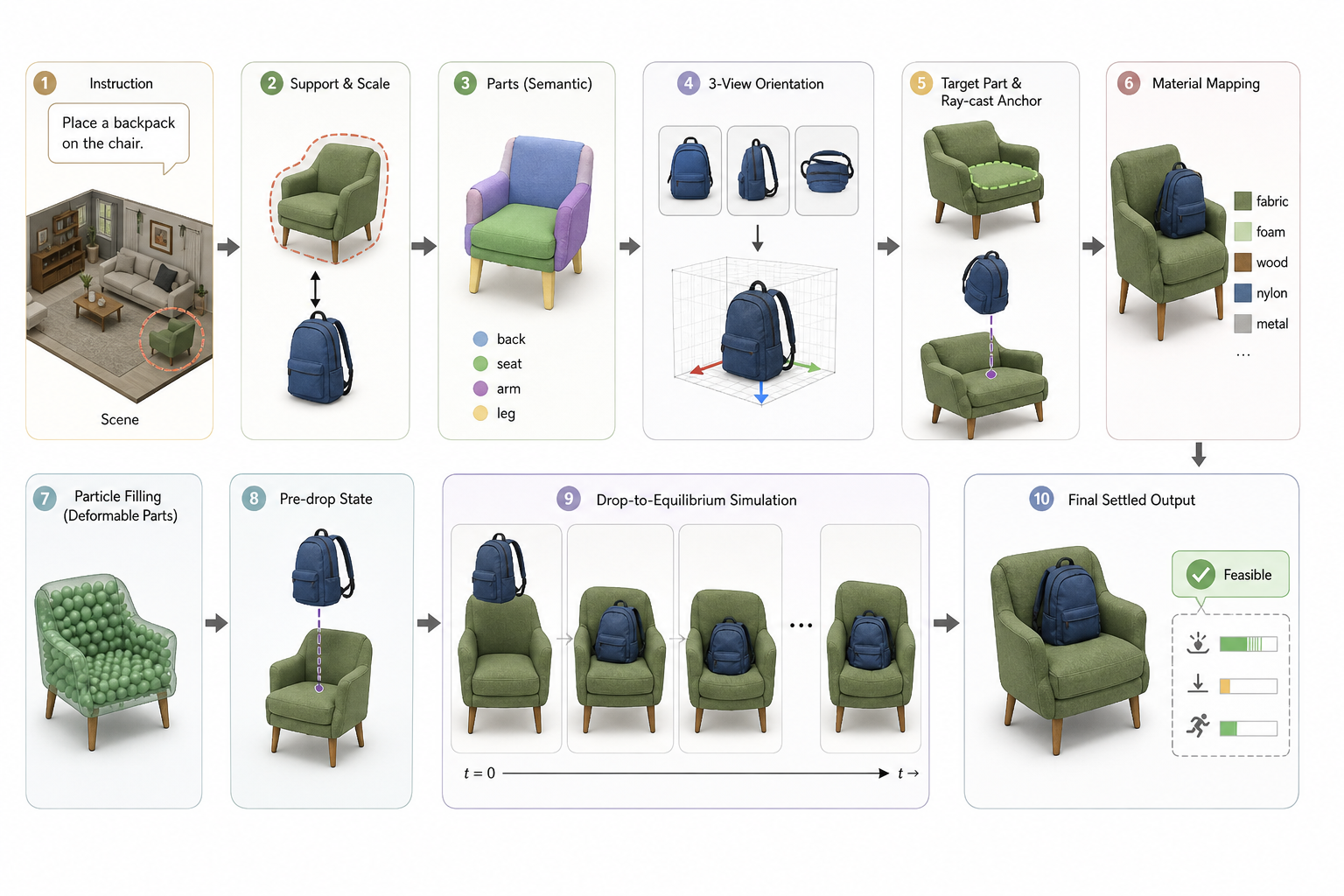}
\captionsetup{type=figure,hypcap=false}
\caption{\method{} overview. The method first builds a release state from the prompt, scene, and object mesh. It gives the simulator a target, a scale, and an orientation. It also defines material roles and the particle representation. A rigid--deformable simulator then settles the object and returns the final state with diagnostics and a feasibility flag.}
\label{fig:pipeline}
\vspace{-0.9em}
\end{center}

\begin{abstract}
Current language-guided 3D object placement typically predicts a rigid 6-DoF pose, ignoring the physical reality that indoor placements are dictated by contact. Consequently, predicted poses often float, interpenetrate supports, or fail to deform soft surfaces. We propose \textbf{D}eformable \textbf{C}ontact-\textbf{A}ware \textbf{P}lacement(DCAP), a framework that reframes placement as a \textit{drop-to-equilibrium} process. Rather than predicting a static pose, DCAP constructs a semantic release state that specifies the drop location, object scale, and deformable components, allowing a coupled rigid- deformable simulator to settle the object under gravity and friction. This explicitly separates semantic intent from physical realization. Evaluated on a benchmark of 2,063 prompt-asset-scene cases, DCAP significantly outperforms static baselines. On standard placements, it improves VLM Physics \& Naturalness by 21.2 points and human preference by 26.5 points over the strongest raw rigid baseline, while reducing mean penetration depth by 72.3\%. Compared to simulated rollout baselines, DCAP reduces residual speed by 72.1\%. Ultimately, our results prove that physically accurate 3D placement requires modeling semantic intent and physical realization as strictly separate processes.
\end{abstract}

\section{Introduction}

Language-guided 3D object placement holds strong promise for embodied AI, synthetic data generation, and virtual scene design, because it lets users arrange environments through natural-language instructions \citep{savva2019habitat,kolve2017ai2thor,yang2024holodeck,li2025freeinsert}. In the physical world, however, placement is rarely a static event. It is a dynamic interaction. Ideally, a placement system should reflect the fact that every placement is fundamentally a drop-to-equilibrium process in which an object settles until gravity, frictional contact, and material response to reach balance \citep{todorov2012mujoco,stomakhin2013mpm,jiang2015apic,hu2018mlsmpm,muller2007pbd}. Such physically grounded, contact-aware placements would provide better foundations for robotic manipulation training, interactive virtual environments, and realistic scene synthesis \citep{savva2019habitat,kolve2017ai2thor,li2025freeinsert,lin2026pat3d}.

Despite impressive progress in 3D-language grounding and scene understanding \citep{chen2020scanrefer,hong2023threedllm,zhu2024scanreason}, current placement models are still primarily designed for rigid, semantic-only pose prediction. Models such as PlaceIt3D and FirePlace study the problem of inserting an asset into a 3D scene by directly predicting a final 6-DoF rigid pose \citep{abdelreheem2025placeit3d,huang2025fireplace}. Yet many indoor supports such as beds, cushions, and upholstered chairs are not rigid shelves. Their deformation changes the final height, contact patch, and visible shape of the placed object \citep{muller2007pbd,stomakhin2013mpm,hu2018mlsmpm}. As a result, static predictions may be physically implausible: objects may hover above fabric, fail to press into soft supports, or interpenetrate the scene geometry, which leads to failure modes that are difficult to diagnose from semantic placement alone \citep{liu2021opa,wu2024gpteval3d}. These limitations make rigid-only placement a poor fit for interactive, physics-driven simulation.

To bridge the gap between semantic, non-deformable placement and physically faithful 3D realization, we argue that the settling process itself must be part of the task rather than a post hoc correction. We introduce \textbf{Deformable Contact-Aware Placement} (\method{}), a framework for physics-aware 3D object insertion, as illustrated in Figure~\ref{fig:pipeline}. Given a reconstructed scene, a prompt, and an object mesh, \method{} does not directly guess the final pose. Instead, it first constructs a semantically grounded release state using language, vision, and geometry cues from modern 3D grounding and open-vocabulary perception pipelines \citep{chen2020scanrefer,kirillov2023segment,cheng2022mask2former,peng2023openscene,jatavallabhula2023conceptfusion,yang2024sampart3d}. This state identifies the support, sets the metric scale and coarse orientation, and assigns material and deformable roles. A coupled rigid--deformable simulator then lets the object settle naturally under gravity and frictional contact \citep{stomakhin2013mpm,jiang2015apic,hu2018mlsmpm,genesis2025a,genesis2025b}. This design separates what the user intended from what the physical world permits, and allows the system to return both a settled state and an interpretable feasibility flag when an instruction is physically impossible.

The separation is useful in three ways. First, compliant supports and deformable objects become natural cases rather than special cases. The final pose is produced by material response, not by a rigid correction. Second, failure becomes a valid output. If the support is too narrow, the object is too large, or the motion does not converge, the system can reject the placement instead of returning an implausible pose. Third, the method exposes physical diagnostics that are difficult to see in a few renderings. They measure penetration, contact, residual motion, and target drift. Such diagnostics complement human and VLM judgments, which are useful but can miss physical errors hidden in geometry or dynamics \citep{wu2024gpteval3d,tam2026sceneeval}.

\textbf{Contributions}: 1) We formulate language-guided 3D object placement as release-state planning followed by drop-to-equilibrium settling; 2) We introduce \method{}, a modular pipeline that combines language/vision grounding with rigid--deformable simulation; 3) We build a benchmark of simulation-ready indoor scenes paired with placement prompts and rigid or deformable assets; 4) We evaluate placement along semantic and physical axes, with objective diagnostics, ablations, submodule checks, and failure analysis.

\section{Related Work}

\textbf{Language-guided 3D placement, grounding, and layout.} Language-guided placement connects a prompt to a support, a region, an orientation, and a set of spatial constraints. PlaceIt3D and FirePlace study this problem directly for asset insertion into real or reconstructed scenes \citep{abdelreheem2025placeit3d,huang2025fireplace}. Broader 3D-language and layout methods use LLMs, VLMs, scene graphs, differentiable optimization, or embodied-scene priors to arrange objects in 3D \citep{hong2023threedllm,zhu2024scanreason,huang2025reason3d,feng2023layoutgpt,yang2024holodeck,sun2025layoutvlm,ran2025directlayout,yang2025optiscene}. These methods are valuable because they provide semantic intent and approximate geometry. \method{} uses the same kind of grounding signal, including replaceable open-vocabulary segmentation backends \citep{chen2020scanrefer,kirillov2023segment,cheng2022mask2former,peng2023openscene,jatavallabhula2023conceptfusion}. The difference is where these signals enter the problem. In \method{}, they define a release state. The final pose is resolved by contact and compliance.

\textbf{Scene resources and visual object insertion.} Large indoor reconstructions and object repositories have enabled progress in scene understanding, embodied navigation, environment generation, and asset insertion \citep{dai2017scannet,chang2017matterport3d,ramakrishnan2021hm3d,xia2018gibson,savva2019habitat,kolve2017ai2thor,deitke2023objaverse,yang2024sampart3d}. These resources are often designed for perception or rendering, not for deformable contact. Image-space and 3DGS insertion methods further improve visual plausibility through appearance matching, shadows, relighting, or multi-view consistency \citep{liu2021opa,zhu2023topnet,li2025freeinsert}, complementary to \method{}. A visually plausible insertion can still be physically invalid, and a settled 3D state can later be rendered with stronger image-space tools. Our benchmark therefore focuses on the requirements of physical placement, including metric scale, usable support geometry, material labels, and contact diagnostics.

\textbf{Physical simulation and diagnostic evaluation.} Frictional contact and deformation are standard topics in MPM, PBD, and rigid-body simulation \citep{stomakhin2013mpm,jiang2015apic,hu2018mlsmpm,muller2007pbd,todorov2012mujoco,makoviychuk2021isaacgym,genesis2025b,genesis2025a,physx2025}. MPM is a natural fit for compliant placement because particles represent soft material while a background grid transfers stress and contact. Rigid-body engines remain efficient for stiff components and mature collision handling. Recent physics-augmented scene generation also uses simulation to improve equilibrium and non-intersection, but it mainly targets rigid scene-level generation rather than deformable insertion into a fixed reconstruction \citep{lin2026pat3d}. \method{} combines these views by simulating soft or contact-critical parts deformably while keeping stiff components rigid. For evaluation, we pair human and VLM judgments with direct measurements of contact, motion, and convergence \citep{wu2024gpteval3d,tam2026sceneeval}.

\section{Problem Setup}
\label{sec:problem_setup}

Given a reconstructed indoor scene mesh \(S\), a natural-language placement instruction \(p\), and an object mesh \(M_O\), our goal is to produce a physically settled placement outcome
\[
    y =
    \left(
    R^\star,
    \mathbf{t}^\star,
    \widetilde{M}_O,
    \widetilde{S},
    f,
    d
    \right).
\]
Here \(R^\star \in SO(3)\), the 3D rotation group, and \(\mathbf{t}^\star \in \mathbb{R}^3\) are the final object rotation and translation in scene coordinates. The tildes denote the settled geometries after pose application and any simulated deformation. The flag \(f\in\{0,1\}\) indicates whether placement is feasible, and \(d\) stores diagnostic measurements such as penetration, support contact, target drift, residual motion, and convergence. The core modeling choice in DCAP is to separate semantic intent from physical realization. DCAP does not directly predict the final pose \((R^\star,\mathbf{t}^\star)\). Instead, it constructs a semantic release state and lets a physics-based drop-to-equilibrium rollout determine the final settled outcome.

\paragraph{Release state.}
\label{sec:predrop_state}
We define the pre-drop release state as
\[
    z_0 =
    \left(
    b,
    s,
    S_b,
    P_b,
    R_0,
    \mathbf{t}_0,
    \pi,
    a,
    \theta,
    \mathcal{R},
    \mathcal{D},
    X_0
    \right).
\]
Here \(b\) is the intended support category, \(S_b\subseteq S\) is the selected support instance, and \(P_b\) is the set of semantic parts of that support. The scalar \(s>0\) is a category-level real-world scale prior for the object; it is not chosen to guarantee that the object fits on the selected support. The pair \((R_0,\mathbf{t}_0)\) is the initial release pose. The symbol \(\pi\) denotes the target support part, such as a cushion or shelf board, and \(a\) denotes the anchor within that part, such as center, left, right, front, back, top, head, or foot. The material map \(\theta\) assigns physical parameters such as density, stiffness, and friction. The sets \(\mathcal{R}\) and \(\mathcal{D}\) contain the components treated as rigid and deformable, respectively, and \(X_0\) stores the initial Material Point Method (MPM) particles for deformable components. A release state is an intended initialization, not a guarantee of physical feasibility. Even if \(z_0\) is semantically correct, the scaled object may be too large for the selected support, the support may not provide stable contact, or the rollout may fail to converge.

\paragraph{Drop-to-equilibrium settling.}
\label{sec:drop_to_equilibrium}
Given \(z_0\), DCAP computes the final outcome through a rigid--deformable rollout
\[
    y = \Gamma(z_0; S, M_O, \mathbf{g}),
\]
where \(\Gamma\) denotes the simulator and diagnostic operator, and \(\mathbf{g}\) is the gravity vector. The rollout applies the scale \(s\), initializes the object at \((R_0,\mathbf{t}_0)\), couples rigid bodies with deformable MPM particles, and simulates contact under gravity and friction. It terminates when the system reaches equilibrium or when the maximum simulation horizon is reached. Equilibrium is detected using velocity, center-of-mass drift, and bounding-box change thresholds over a fixed convergence window. Appendix~\ref{app:simulation} gives thresholds and simulator settings. We avoid treating rollout as post-processing correction: its output is the physical realization of the release state.

\paragraph{Feasibility and diagnostics.}
\label{sec:feasibility_diagnostics}
A placement is marked feasible only if three requirements are satisfied: the rollout converges, the object remains on the intended support \(S_b\), and the diagnostic measurements in \(d\) stay within tolerance. The diagnostics include mean penetration depth, support contact ratio, target drift from the intended part-anchor pair \((\pi,a)\), residual speed, and convergence status. Appendix~\ref{app:metrics} defines these diagnostics, and Appendix~\ref{app:simulation} lists the numerical tolerances. This definition separates different failure modes. A placement can be dynamically invalid if it does not settle, semantically off target if it settles on the wrong support or region, or infeasible if it violates any required semantic, contact, convergence, or tolerance condition. Static baselines do not run the rollout operator \(\Gamma\), so dynamic quantities such as residual speed and convergence are undefined for them. They can still be evaluated using static geometric and semantic diagnostics, such as support correctness, penetration, contact, and target drift.

\section{Method: \method}
\label{sec:method}

\begin{figure}[t]
\centering
\includegraphics[width=0.96\linewidth]{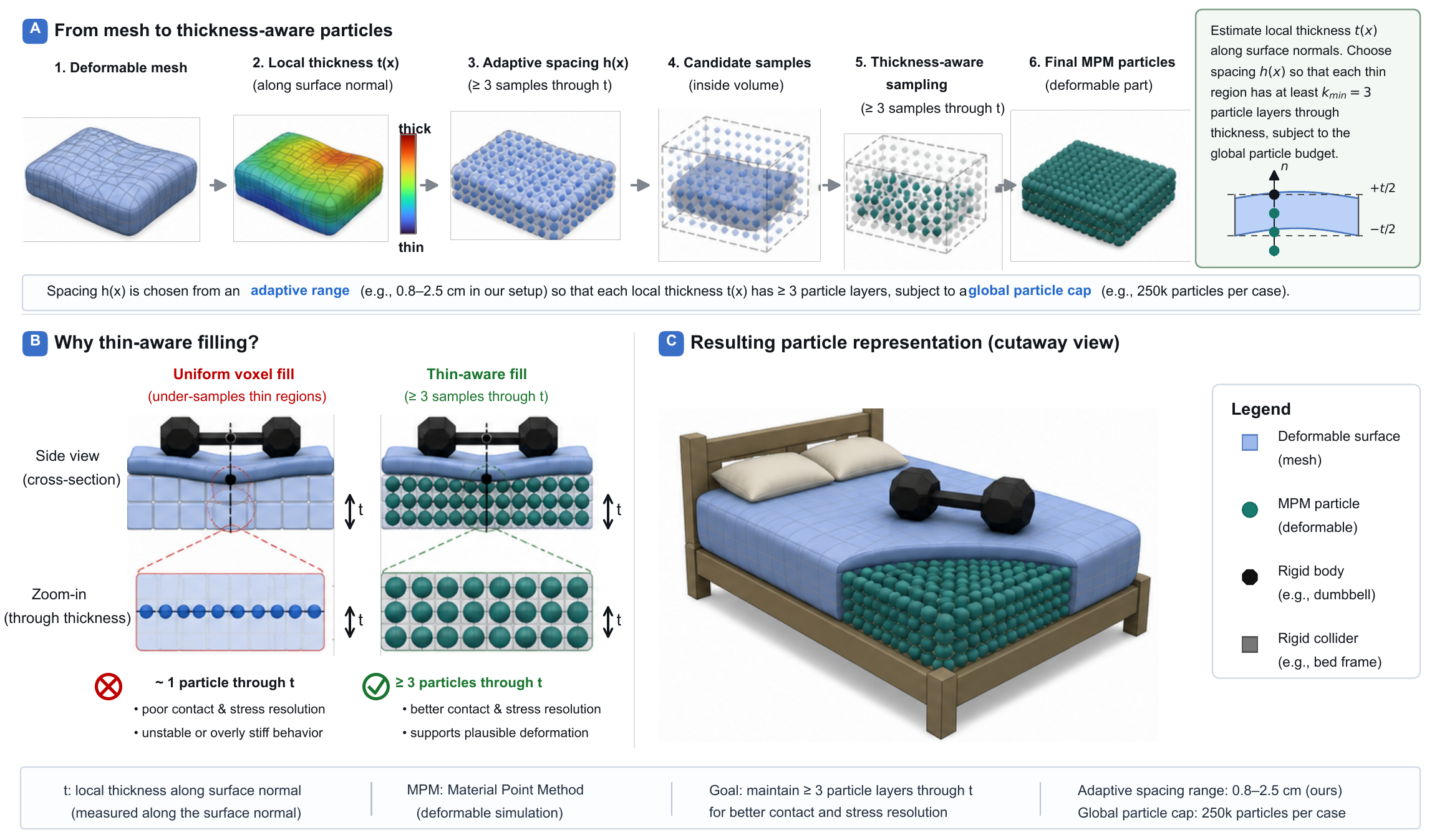}
\caption{\textbf{Thin-aware particle filling.} \method{} estimates local thickness and adapts particle spacing so thin compliant regions retain enough through-volume samples for contact and stress resolution.}
\label{fig:thin-aware-filling}
\vspace{-0.5em}
\end{figure}

\method{} is modular. As shown in Figure~\ref{fig:pipeline}, it separates semantic release-state construction from physical realization. DCAP maps the inputs to a release state through a planner:
\begin{equation}
    A_\psi : (S,p,M_O) \mapsto z_0,
    \qquad
    y = \Gamma(z_0; S, M_O, \mathbf{g}).
\end{equation}
Here \(A_\psi\) denotes DCAP's planner, and \(\psi\) collects its fixed prompts, perception modules, and geometric rules. Upgrading any module, such as the LLM, VLM, segmenter, or simulator, improves one estimator inside \(A_\psi\) without changing the underlying task definition. Each stage estimates part of \(z_0\), records uncertainty when needed, and passes a physically meaningful state to the simulator. Table~\ref{tab:method_variables} summarizes how  modules instantiate the release-state, and Figure~\ref{fig:thin-aware-filling} shows particle-filling; Appendices~\ref{app:prompts}, \ref{app:segmentation}, and~\ref{app:materials} give the prompt schemas, segmentation details, and material library.

\begin{table}[!htb]
\centering
\caption{\method{} modules for release-state construction and physical rollout.}
\label{tab:method_variables}
\small
\setlength{\tabcolsep}{4.5pt}
\begin{tabularx}{\linewidth}{llX}
\toprule
Component & Module & Role \\
\midrule
\(b,s\) & Support grounding and scale prior & Infer the intended support category and a physically meaningful object scale \\
\(S_b,P_b\) & Segmentation and parts & Identify the support instance and decompose it into usable semantic parts \\
\(R_0\) & Three-candidate VLM orientation & Select a coarse release orientation from prompt-conditioned rendered candidates \\
\(\pi,a,\mathbf{t}_0\) & Ray-cast anchoring & Convert the prompt into a release point above the target region \\
\(\theta,\mathcal{R},\mathcal{D}\) & Materials and split & Assign discrete material values and rigid/deformable roles \\
\(X_0\) & Thin-aware particle filling & Initialize a contact-ready deformable particle set \\
\(y\) & Drop-to-equilibrium simulation & Rollout output; not a component of \(z_0\) \\
\bottomrule
\end{tabularx}
\end{table}

\subsection{Semantic Grounding and Scene Decomposition}
\label{sec:semantic_grounding}

Because physics-based simulation is acutely sensitive to metric scale and boundary constraints, \method{} first establishes a formal mapping between semantic intent and the reconstructed scene. We leverage the common-sense reasoning of a large language model (LLM) to extract the intended support category \(b\) and a realistic longest-dimension prior from the instruction \(p\). By isotropically rescaling the object mesh to match this category-level size prior, we obtain metric collision geometry and an approximate volume scale for simulation. Mass and inertia are then approximated from the scaled geometry together with the material parameters in \(\theta\). Given the identified support category \(b\), the system then locates the specific support instance \(S_b\) and decomposes it into semantic parts \(P_b\) (Figure~\ref{fig:pipeline}, Steps~1--3). We follow a standard 2D-to-3D open-vocabulary pipeline to segment the scene into semantic parts (e.g., mattress vs. frame) \citep{kirillov2023segment,cheng2022mask2former,peng2023openscene,jatavallabhula2023conceptfusion}. Beyond visually grounding the prompt, this decomposition establishes material properties to restrict expensive deformable simulation only to necessary regions.

\subsection{Release Pose Initialization}
\label{sec:release_pose}

With the object scaled and the scene segmented, the next step is to define the release pose \((R_0,\mathbf{t}_0)\). Natural language often provides only coarse orientation cues, such as a mug handle facing outward. To efficiently resolve this angular ambiguity, \method{} renders three prompt-conditioned orientation candidates created by 180-degree flips about canonical object axes and uses a vision-language model (VLM) to select the best semantic match (Figure~\ref{fig:pipeline}, Step~4). This stage resolves only coarse semantic orientation, and the rollout handles small contact-induced pose adjustments rather than performing a continuous semantic orientation search. To determine the precise release translation \(\mathbf{t}_0\), we utilize a geometric ray-cast anchoring mechanism. The system maps the prompt to a target part \(\pi\) and an anchor \(a\) and then casts a ray along the gravity direction to identify the corresponding physical point on the support surface (Figure~\ref{fig:pipeline}, Step~5). To ensure numerical stability and provide sufficient headroom for gravitational settling, the object is initialized with a small positive vertical clearance $\delta>0$ (specified in Appendix~\ref{app:simulation}) above the target surface. This guarantees that the settling process is driven entirely by physical mechanics rather than artificial forces from initial mesh interpenetration.

\subsection{Material Assignment and Particle Discretization}
\label{sec:material_discretization}

To balance simulation fidelity with computational efficiency, \method{} must assign physical roles to the parsed geometric parts. Instead of learning continuous material parameters, we map each semantic part to a discrete material ID from a curated library, which provides the density, elastic modulus, and friction parameters collected in \(\theta\) (Figure~\ref{fig:pipeline}, Step~6). Compliant materials (foam, fabric, mattresses) are assigned to $\mathcal{D}$ and stiff materials (wood, metal, frames) are assigned to $\mathcal{R}$. Simulating only compliant parts (e.g., cushions) deformably while keeping stiff frames rigid drastically reduces compute without sacrificing contact fidelity. Once compliant parts are identified, they must be discretized into MPM particles (Figure~\ref{fig:pipeline}, Step~7). Uniform voxel filling often catastrophically under-samples thin indoor surfaces, such as blankets, rendering the geometry numerically fragile and incapable of resolving frictional contact. As illustrated in Figure~\ref{fig:thin-aware-filling}, for a surface point \(\mathbf{x}\), \method{} estimates the local thickness \(t(\mathbf{x})\) along the surface normal and adapts the particle spacing \(h(\mathbf{x})\). The constant \(k_{\min}=3\) is the required minimum number of particle layers through the local thickness. This targeted discretization, bounded by a global particle cap, ensures even thin compliant regions can realistically carry stress and deform naturally.

\subsection{Drop-to-Equilibrium Simulation and Diagnostics}
\label{sec:simulation_method}

The final stage is where placement becomes physics. The object is released from \((R_0,\mathbf{t}_0)\), allowing rigid bodies and MPM particles to interact through frictional contact until the strict convergence criteria from Section~\ref{sec:drop_to_equilibrium} are met or the maximum horizon is reached (Figure~\ref{fig:pipeline}, Steps~8--10). The rollout outputs the final transform, the deformed scene geometry, and an array of physical diagnostics including residual motion, contact ratios, and a feasibility flag. Crucially, rather than collapsing uncertainty into an opaque score, \method{} propagates module-level confidence and diagnostic flags. Low-confidence or invalid module outputs trigger deterministic fallback rules before simulation such as scale clamping, canonical stable-orientation selection, default material assignment, and particle-budget coarsening. These fallback events are recorded as diagnostic flags; Appendix~\ref{app:submodule_eval} lists the exact triggers and recovery rules. If the object leaves the target support, fails to converge, or violates tolerances, the system marks the case as infeasible and explicitly categorizes the error source (perception, geometry, materials, or physics).

\section{Experiments and Analysis}
\label{sec:experiments}

\subsection{Experimental Settings}
\label{sec:experimental_settings}

\paragraph{A Simulation-Ready Benchmark.}
Unlike existing semantic-only placement benchmarks such as PlaceIt3D \citep{abdelreheem2025placeit3d}, evaluating \method{} requires physics-ready geometry. We construct a benchmark of 2,063 prompt--asset--scene cases drawn from 186 high-fidelity indoor reconstructions and 155 object assets \citep{dai2017scannet,chang2017matterport3d,ramakrishnan2021hm3d,xia2018gibson,deitke2023objaverse,yang2024sampart3d}. To support rigorous contact queries and physical rollouts, we process scenes via TSDF fusion and screened-Poisson meshing to guarantee the watertight, metric surfaces required for physical simulation \citep{curless1996volumetric,kazhdan2006poisson,kazhdan2013screened}. This pipeline preserves metric accuracy and provides watertight surfaces suitable for simulation. We evaluate all methods on two distinct splits. The standard split contains 1,923 cases and serves as the primary testbed for aggregate placement quality, human/VLM preference, and physical diagnostics. It is further diversified by contact regime, with 869 soft-contact, 550 mixed-contact, and 504 rigid-contact cases. The stress split contains 140 cases and is used specifically to evaluate robustness and feasibility handling. It includes 64 out-of-distribution feasible prompts and 76 annotated physically impossible prompts. Correct rejections on this split count only toward feasibility metrics, while any returned feasible placements receive the standard diagnostics. Appendix~\ref{app:benchmark} documents the construction log and reporting denominators.

\paragraph{Baselines and Controlled Variants.}
To isolate the impact of our physics-aware formulation, we benchmark against both existing state-of-the-art models and internally controlled variants. To ensure a fair comparison, all methods receive the same prompt, object scale, scene mesh, and camera views. The raw static baselines are FirePlace, PlaceIt3D, LayoutGPT, and Holodeck \citep{huang2025fireplace,abdelreheem2025placeit3d,feng2023layoutgpt,yang2024holodeck}. Because these models output purely static poses, dynamic metrics such as residual speed and convergence are undefined for them. We also evaluate controlled physics variants to untangle the value of the initial release state from the physical settling process. FirePlace + rigid and FirePlace + deformable use FirePlace's static prediction as the initialization for rigid or deformable settling. Conversely, \method{} static (init+snap) tests the upstream language--vision--geometry planner as a static method by replacing the physics simulator with a naive gravity-axis snap. This geometric snap translates the object only along the gravity axis until it reaches the closest non-penetrating contact with the selected support. It does not simulate in a physics engine. Finally, \method{} rigid-only runs the full pipeline while treating all materials as rigid during rollout. Appendix~\ref{app:baselines} gives baseline adaptation details.

\paragraph{Evaluation Metrics.}
We assess placement success along two complementary dimensions, subjective visual plausibility and objective physical stability. Human evaluators and the VLM compare results along two axes. Right Place (RP) measures semantic accuracy by asking whether the object is on the correct support, in the intended region, and oriented consistently with the prompt. Physics \& Naturalness (P/N) measures physical plausibility, so a high score requires stable support, realistic contact, material-consistent deformation, and absence of floating or deep interpenetration. Bradley--Terry--Luce estimates are derived from 2,520 preference trials with 42 retained participants, with 420 attention-check trials excluded from aggregation. Our objective physical diagnostics explicitly measure mesh penetration depth, not valid soft-support indentation, support contact ratio, residual motion, target adherence, and simulator convergence. Physical diagnostics are reported only for methods that run an explicit rollout. Appendices~\ref{app:human}, \ref{app:vlm}, and~\ref{app:metrics} provide the full evaluation protocols.

\subsection{Main Comparison and Controlled Variants}
\label{sec:comparison_existing_methods}

\begin{figure}[!htbp]
\centering
\includegraphics[width=0.72\linewidth]{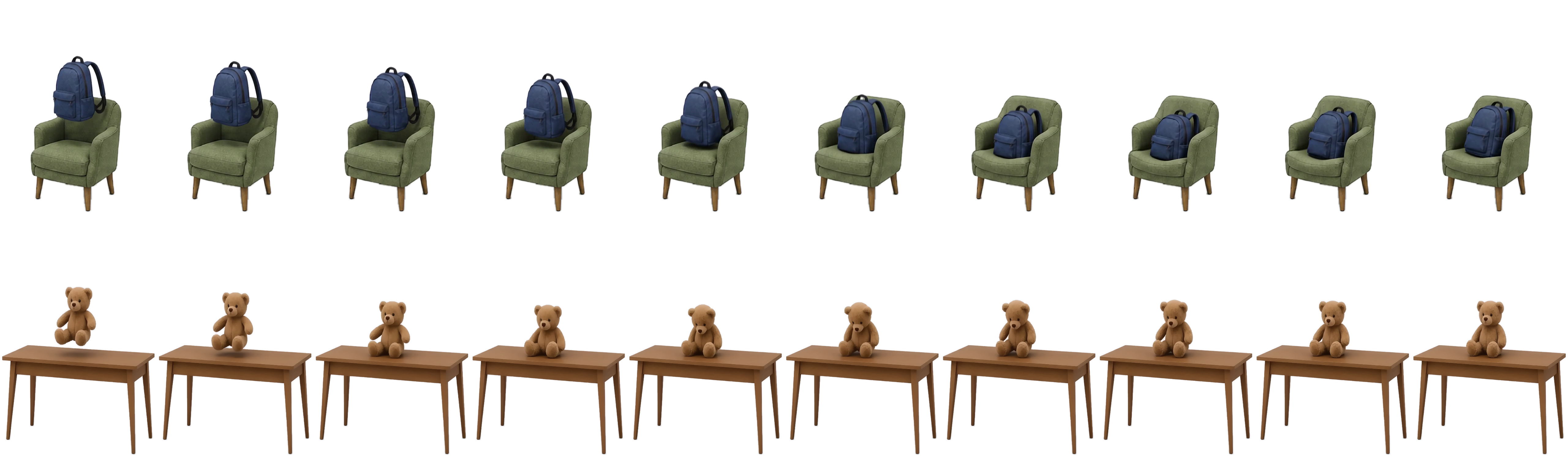}
\caption{Representative settling rollouts. A backpack and a plush toy settle under gravity while preserving the intended support and scale.}
\label{fig:drop_rollouts}
\vspace{-0.45em}
\end{figure}

\begin{figure}[!htbp]
\centering
\includegraphics[width=1\linewidth]{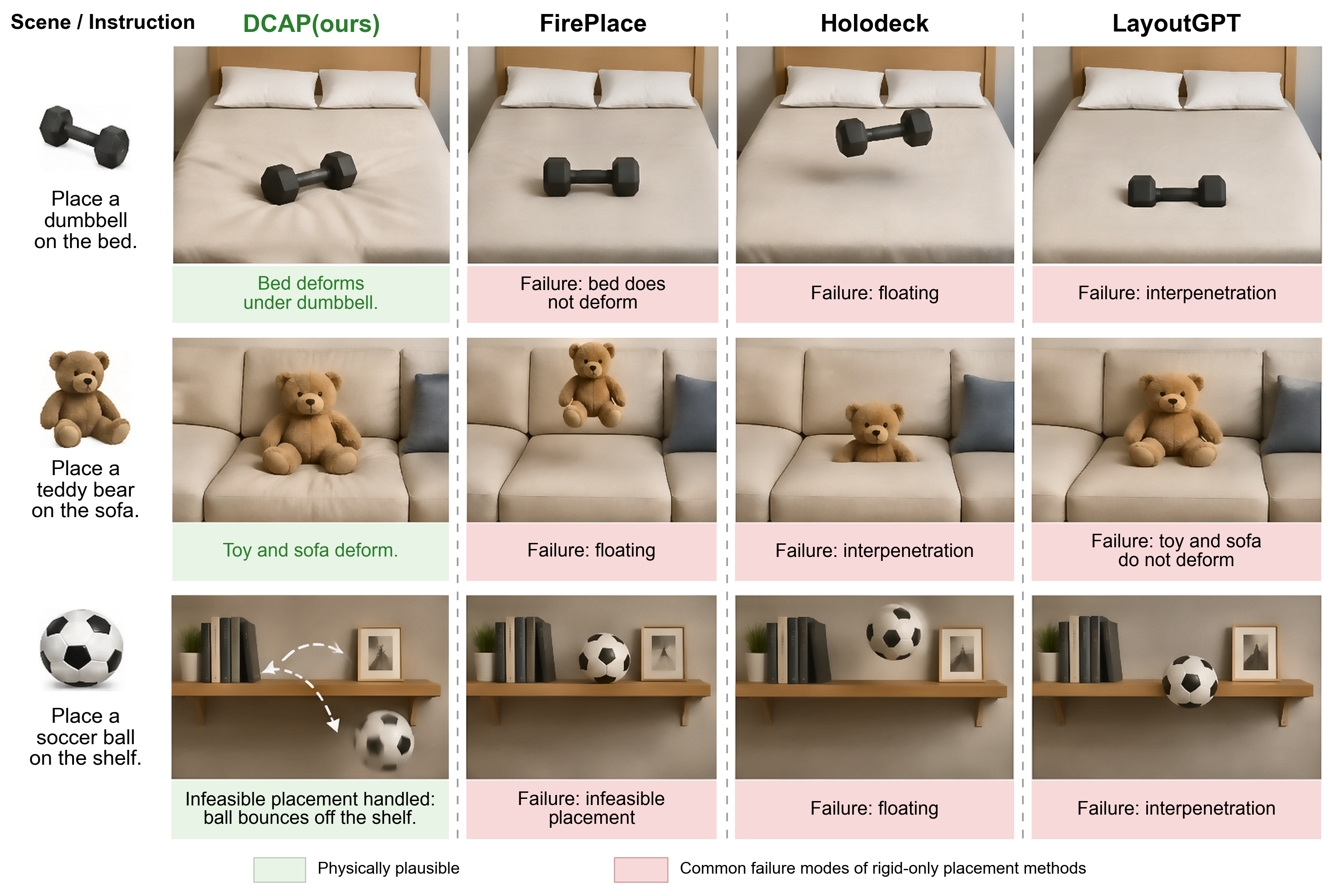}
\caption{Qualitative comparison. \method{} produces stable contact, deformation, and infeasibility handling, while rigid baselines often float, interpenetrate, or miss soft-support deformation.}
\label{fig:qual_drop_rad_style}
\vspace{-0.25em}
\end{figure}

Table~\ref{tab:main_hybrid_rad_style} gives the main comparison, with qualitative examples in Figure~\ref{fig:qual_drop_rad_style} and representative settling sequences in Figure~\ref{fig:drop_rollouts}. \method{} significantly outperforms existing static baselines across all placement and physical metrics. Compared with FirePlace, the strongest raw static baseline in our setting, \method{} raises the VLM Physics \& Naturalness (P/N) score from 64.0 to 85.2 and the human P/N preference score from 50.1 to 76.6. Crucially, it reduces the mean penetration depth from 1.48 cm to 0.41 cm and increases valid support contact from 0.65 to 0.89. To understand where these gains come from, we isolate the two core components of our pipeline, the initial release state and the physical settling process. The controlled variants reveal that both precise initialization and dynamic physical rollout are strictly necessary. The first comparison shows that physical simulation cannot fix a bad initialization. Adding a settling process to the FirePlace baseline improves its performance, and deformable settling helps more than rigid settling. However, FirePlace + deformable still significantly underperforms full \method{}, which isolates the value of our release state. A physics simulator cannot correct an object that is dropped on the wrong part of a surface, oriented incorrectly, or scaled to an implausible size. The reverse comparison makes the complementary point. When we use \method{}'s strong initialization but replace the physics engine with a simple geometric snap (\method{} static), the result is highly competitive with raw static baselines and drastically reduces penetration. Yet its contact ratio and physical naturalness remain limited. A geometric snap can close a gap, but it cannot create a broad contact patch, simulate frictional resting, or deform a soft support. Together, these results explain why we evaluate semantic placement, Right Place (RP), and physical plausibility, Physics \& Naturalness (P/N), as separate metrics. Correct spatial coordinates do not guarantee a correct physical state. A method can place a dumbbell at the center of a mattress and still fail if the mattress does not compress, or place a plush toy on the correct sofa cushion and still fail if it floats. Ultimately, these are not just different target coordinates; they are different physical states.

\begin{table}[!htb]
\centering
\caption{\textbf{Baseline comparison on the standard split (1,923 cases).} Left: aggregate placement quality. Right: controlled ablations; VLM P/N gains report \(B-A\) for each \(A\!\to\!B\) contrast, with ``ref.'' rows defining block references.}
\label{tab:main_hybrid_rad_style}
{\scriptsize
\setlength{\tabcolsep}{2.1pt}
\renewcommand{\arraystretch}{0.91}
\resizebox{\linewidth}{!}{%
\begin{tabular}{@{}lrrrrrr@{\hspace{1.05em}}llrrrr@{}}
\toprule
\multicolumn{7}{c}{Aggregate quality} & \multicolumn{6}{c}{Controlled physics view}\\
\cmidrule(lr){1-7}\cmidrule(l){8-13}
Method & VLM & Human & Pen. & Contact & Spd. & Conv. & Contrast & Change & VLM P/N gain & Pen. drop & Final conv. & Final time\\
& P/N\metricup & P/N\metricup & cm\metricdown & \metricup & cm/s\metricdown & \metricup & & & \metricup & cm\metricup & \metricup & s\metricdown\\
\midrule
FirePlace raw & 64.0 & 50.1 & 1.48 & 0.65 & \na & \na & FirePlace raw & static ref. & ref. & ref. & \na & 14.6\\
PlaceIt3D raw & 60.6 & \na & 1.70 & 0.62 & \na & \na & PlaceIt3D raw & static ref. & ref. & ref. & \na & \na\\
LayoutGPT raw & 49.7 & \na & 2.31 & 0.47 & \na & \na & FP raw$\to$+rigid & rigid rollout & +2.7 & 0.38 & 0.61 & 38.2\\
Holodeck raw & 54.2 & \na & 2.05 & 0.53 & \na & \na & FP raw$\to$+def. & deform. rollout & +8.7 & 0.73 & 0.71 & 86.9\\
FP + rigid & 66.7 & \na & 1.10 & 0.70 & 6.8 & 0.61 & \method{} static & geom. snap & ref. & ref. & \na & 44.1\\
FP + def. & 72.7 & 63.6 & 0.75 & 0.79 & 4.1 & 0.71 & static$\to$rigid & rigid rollout & +11.1 & 0.26 & 0.76 & 69.4\\
\method{} static (init+snap) & 65.9 & \na & 0.86 & 0.58 & \na & \na & rigid$\to$full & add deform. & +8.2 & 0.19 & 0.85 & 107.5\\
\method{} rigid & 77.0 & 68.0 & 0.60 & 0.83 & 3.7 & 0.76 & static$\to$full & full rollout & +19.3 & 0.45 & 0.85 & 107.5\\
\textbf{\method{} full} & \textbf{85.2} & \textbf{76.6} & \textbf{0.41} & \textbf{0.89} & \textbf{1.9} & \textbf{0.85} & FP+def.$\to$full & better init. & +12.5 & 0.34 & 0.85 & 107.5\\
\bottomrule
\end{tabular}}}
\end{table}

\FloatBarrier

\subsection{Regime, Feasibility, and Human Validation}
\label{sec:regime_human_validation}

The strongest gains occur on soft supports like beds and sofas, where \method{} achieves an 88.2 P/N score compared to 60.4 for raw FirePlace and 74.7 for rigid-only \method{} (Tables~\ref{tab:human_diag_rad_style} and~\ref{tab:regime_feas_rad_style}). On these compliant surfaces, final height and contact are dictated by mass, stiffness, and friction rather than original mesh geometry alone. Rigid supports provide a natural control. Here, full \method{} and its rigid-only variant perform similarly, though the full pipeline remains robust due to strict convergence and feasibility checks. Beyond spatial accuracy, the stress split proves \method{} can reliably reject physically invalid prompts (0.83 F1) while accepting 52 of 64 feasible out-of-distribution cases. Human evaluations echo this physical robustness, showing a 26.5-point BTL preference over raw FirePlace in physical naturalness. Crucially, comparing full and rigid-only \method{} reveals an 8.6-point gap in naturalness (P/N) but only a 0.7-point gap in spatial accuracy (RP). This proves that deformable settling enhances realism without displacing the object from its semantic target. Ultimately, the strong alignment across human preference, VLM scoring, and objective diagnostics confirms that our gains are physically grounded rather than merely cosmetic.

\begin{table}[!htb]
\centering
\caption{\textbf{Human preference and objective diagnostics.} Bradley--Terry--Luce (BTL) estimates from forced-choice evaluations strongly align with automated objective metrics. Asterisks indicate that the paired bootstrap confidence interval for the human BTL difference excludes zero after Holm correction; significance markers apply only to human forced-choice differences.}
\label{tab:human_diag_rad_style}
{\scriptsize
\setlength{\tabcolsep}{2.6pt}
\renewcommand{\arraystretch}{0.94}
\resizebox{\linewidth}{!}{%
\begin{tabular}{@{}llrrr@{\hspace{1.7em}}lrrrr@{}}
\toprule
\multicolumn{5}{c}{Human forced-choice results} & \multicolumn{5}{c}{Objective diagnostics}\\
\cmidrule(lr){1-5}\cmidrule(l){6-10}
Comparison & Axis & \method{} & Baseline & Diff. & Method & Pen. & Contact & Speed & Conv.\\
& & & & & & cm\metricdown & \metricup & cm/s\metricdown & \metricup\\
\midrule
\method{} vs FP raw & RP & 61.2 & 56.4 & +4.8* & FirePlace raw & 1.48 & 0.65 & \na & \na\\
\method{} vs FP raw & P/N & 76.6 & 50.1 & +26.5* & FP+rigid & 1.10 & 0.70 & 6.8 & 0.61\\
\method{} vs FP+def. & RP & 61.2 & 55.6 & +5.6* & FP+def. & 0.75 & 0.79 & 4.1 & 0.71\\
\method{} vs FP+def. & P/N & 76.6 & 63.6 & +13.0* & \method{} static & 0.86 & 0.58 & \na & \na\\
\method{} vs \method{} rigid & RP & 61.2 & 60.5 & +0.7 & \method{} rigid & 0.60 & 0.83 & 3.7 & 0.76\\
\method{} vs \method{} rigid & P/N & 76.6 & 68.0 & +8.6* & \textbf{\method{} full} & \textbf{0.41} & \textbf{0.89} & \textbf{1.9} & \textbf{0.85}\\
\bottomrule
\end{tabular}}}
\end{table}

\begin{table}[!htb]
\centering
\caption{\textbf{Performance across contact regimes and stress feasibility.} Left: \method{} outperforms baselines on soft supports. Bold indicates best P/N score within each contact regime; convergence is reported after the slash. Right: robustness on extreme stress cases, measured by discrete rejection/acceptance counts rather than continuous diagnostics.}
\label{tab:regime_feas_rad_style}
{\scriptsize
\setlength{\tabcolsep}{2.6pt}
\renewcommand{\arraystretch}{0.94}
\resizebox{\linewidth}{!}{%
\begin{tabular}{@{}lrrrrr@{\hspace{1.6em}}lrrr@{}}
\toprule
\multicolumn{6}{c}{Contact-regime placement quality} & \multicolumn{4}{c}{Stress feasibility}\\
\cmidrule(lr){1-6}\cmidrule(l){7-10}
Regime & Cases & FP raw & FP+def. & \method{} rigid & \method{} full & Stress statistic & Infeas. & Feas. & Summary\\
& & P/N & P/N & P/N & P/N / Conv. & & & &\\
\midrule
Soft & 869 & 60.4 & 70.4 & 74.7 & \textbf{88.2} / 0.89 & Pred. reject & 62 & 12 & 74\\
Mixed & 550 & 64.8 & 74.0 & 78.0 & \textbf{84.5} / 0.84 & Pred. feasible & 14 & 52 & 66\\
Rigid & 504 & 69.2 & 75.2 & 80.0 & \textbf{80.9} / 0.80 & Reject P/R/F1 & 0.84 & 0.82 & 0.83\\
Standard avg. & 1,923 & 64.0 & 72.7 & 77.0 & \textbf{85.2} / 0.85 & OOD true accept & -- & 0.81 & 52/64\\
\bottomrule
\end{tabular}}}
\end{table}

\FloatBarrier

\subsection{Ablations, Runtime, and Failure Modes}
\label{sec:ablations}

Table~\ref{tab:abl_runtime_rad_style} shows that robust physical simulation strictly depends on accurate upstream data. Degrading geometric fidelity by removing thin-aware filling drops the P/N score to 76.3 because coarse particles cannot support stable contact. Material precision is equally critical. Substituting generic defaults (78.0 P/N) or rigid supports (74.5 P/N) severely cripples the ability to model compliant deformation. Furthermore, physics engines cannot fix wrong semantic targets. Our fallback rules are therefore essential for mitigating inevitable perception errors and preventing hard failures from spiking (0.184 vs. 0.039). Given the 107.5-second average runtime bottlenecked by simulation, this performance profile points to a practical cascade that we can deploy fast rigid placement for stiff scenes and reserve expensive deformable settling for soft materials or uncertain contacts. Appendices~\ref{app:submodule}--\ref{app:videos} detail further submodule checks and rollouts.

\begin{table}[!ht]
\centering
\caption{\textbf{System ablations, runtime, and sensitivity.} Left: Module-level ablations evaluated via VLM and objective metrics. Right: Pipeline runtime breakdown and hyperparameter sensitivity. Note: The left and right blocks are independent and not row-paired.}
\label{tab:abl_runtime_rad_style}
{\scriptsize
\setlength{\tabcolsep}{2.6pt}
\renewcommand{\arraystretch}{0.92}
\resizebox{\linewidth}{!}{%
\begin{tabular}{@{}lrrrr@{\hspace{1.7em}}lrl@{}}
\toprule
\multicolumn{5}{c}{Ablation and sensitivity results} & \multicolumn{3}{c}{Runtime and sensitivity}\\
\cmidrule(lr){1-5}\cmidrule(l){6-8}
Variant / setting & P/N & Pen. & Conv. & Hard fail & Stage / setting & Value & Note\\
\midrule
\method{} full & \textbf{84.1} & \textbf{0.42} & \textbf{0.84} & \textbf{0.039} & Full pipeline & 107.5 s & total mean\\
No thin-aware fill & 76.3 & 0.58 & 0.75 & 0.094 & Median pipeline & 88.4 s & median\\
No material-specific map & 78.0 & 0.55 & 0.77 & 0.081 & Simulation & 54.2 s & main bottleneck\\
Rigid support only & 74.5 & 0.64 & 0.72 & 0.103 & Segmentation + parts & 18.6 s & scene processing\\
Rigid object only & 79.0 & 0.57 & 0.78 & 0.073 & Thin-aware filling & 12.8 s & particle count\\
No target sub-part & 76.8 & 0.63 & 0.74 & 0.105 & Orientation VLM & 8.0 s & 3 candidates\\
No fallback rules & 78.0 & 0.62 & 0.76 & 0.184 & Target/overhead & 2.7 s & anchoring/retries\\
Coarser particles & 79.8 & -- & -- & -- & Coarser particles & 79.8 P/N & faster, less stable\\
Finer particles & 84.8 & -- & -- & -- & Finer particles & 84.8 P/N & 156.7 s total\\
Shorter horizon & 80.7 & -- & -- & -- & Shorter horizon & 80.7 P/N & under-settled\\
Longer horizon & 84.4 & -- & -- & -- & Longer horizon & 84.4 P/N & 132.4 s total\\
\bottomrule
\end{tabular}}}
\end{table}

\FloatBarrier

\section{Discussion and Conclusion}

While DCAP establishes 3D placement as a physical \textit{drop-to-equilibrium} process, its robustness is inherently bounded by current simulation constraints. Our material library relies on simplified isotropic parameters, and placement fidelity is strictly bottlenecked by the quality of upstream geometry reconstruction. Furthermore, while humans and VLMs excel at assessing visual plausibility, they often overlook subtle physical violations that only objective metrics capture. Bridging these gaps will require future benchmarks that incorporate real-world physical validation for complex, deformable household materials. Despite these limitations, our coupled rigid-deformable pipeline significantly outperforms static baselines, demonstrating that realistic resting states cannot be recovered by 6-DoF pose estimation alone. Ultimately, this work highlights a fundamental truth for embodied scene synthesis. Assigned spatial coordinates are not a substitute for physical realization, and being in the right place is never the same as being physically placed.

\clearpage
\appendix
\section*{Appendix}
\addcontentsline{toc}{section}{Appendix}

\section{Implementation and Reproducibility Details}
\label{app:implementation}

Table~\ref{tab:implementation_config} fixes the concrete implementation used for the reported experiments. The method is modular, but all numbers in the main paper use this configuration unless explicitly marked as an ablation. Cloud model calls are run with deterministic decoding when supported; invalid JSON responses are retried with the same prompt plus a parser error message before any geometric fallback is applied.

\begin{table}[ht]
\centering
\caption{Concrete implementation settings for the reported experiments.}
\label{tab:implementation_config}
\small
\setlength{\tabcolsep}{4pt}
\begin{tabularx}{\linewidth}{P{0.25\linewidth}Y}
\toprule
Component & Setting \\
\midrule
LLM parsing, target-part selection, material mapping & \texttt{gpt-5.2}; reasoning effort low; max output 512 tokens; structured JSON output; seed 0 where supported; two schema-repair retries before fallback \\
Orientation VLM inside \method{} & \texttt{gpt-5.2}; reasoning effort low; max output 256 tokens; three rendered candidates; no access to final evaluation scores \\
VLM evaluator & \texttt{gemini-3-pro-preview}; temperature 0; top-p 1.0; max output 512 tokens; randomized method order; separate prompts from the orientation module \\
Segmentation backend & OpenScene mesh features, Segment Anything ViT-H masks, and Mask2Former panoptic labels fused from 48 RGB-D render views; fallback to geometric support extraction when confidence is low \\
Renderer & Blender 4.2.3 with Eevee Next rasterization; 1024$\times$1024 PNGs for VLM scoring and 768$\times$768 panels for human trials; fixed focal length, exposure, and environment light across methods \\
Simulator & Genesis 0.2.1-compatible rigid--MPM backend with custom two-way frictional contact; Python 3.11, CUDA 12.4, single-precision GPU kernels \\
Runtime hardware & One NVIDIA RTX 4090 GPU, 24 GB VRAM; AMD Threadripper Pro 5975WX CPU; 256 GB RAM; Ubuntu 22.04; timings include rendering, segmentation, model-call latency, particle filling, simulation, and diagnostics, but exclude one-time mesh reconstruction \\
\bottomrule
\end{tabularx}
\end{table}

\section{Prompts and Model Calls}\label{app:prompts}

This appendix records the prompt templates and constrained output schemas used by the LLM and VLM components. All LLM calls use the parser settings in Table~\ref{tab:implementation_config}. Confidence below 0.6 triggers a constrained retry or the fallback listed in Table~\ref{tab:fallback_rules}.

\subsection{LLM: Intent Parsing and Scale Prior}

\textbf{System prompt.} You are a careful assistant for a 3D placement system. Output only valid JSON. Use only allowed categories. If uncertain, set confidence below 0.6.

\textbf{User prompt template.}
\begin{quote}\small
Given a natural-language instruction for placing an object into an indoor scene, extract the support category, a realistic longest object dimension in meters, and a confidence value. The support category must come from \texttt{ALLOWED\_BASES}. Return JSON in the following schema: \texttt{\{base: ..., size\_m: ..., confidence: ...\}}. Instruction: \texttt{<PROMPT>}. Object category: \texttt{<OBJ\_CAT>}.
\end{quote}

Table~\ref{tab:size_ranges} lists the size ranges used as sanity checks for common object categories.

\begin{table}[ht]
\centering
\caption{Size sanity ranges.}
\label{tab:size_ranges}
\begin{tabularx}{0.75\linewidth}{lXX}
\toprule
Object category & Min (m) & Max (m) \\
\midrule
Ball & 0.04 & 0.30 \\
Pillow & 0.30 & 0.80 \\
Plush toy & 0.12 & 0.65 \\
Towel & 0.35 & 1.60 \\
Shirt / clothing & 0.45 & 1.20 \\
Dumbbell & 0.20 & 0.55 \\
Book & 0.18 & 0.42 \\
Other / unknown & 0.08 & 1.00 \\
\bottomrule
\end{tabularx}
\end{table}

\subsection{LLM: Target Sub-Part and Anchor}

\textbf{User prompt template.}
\begin{quote}\small
You are given a placement instruction and the available parts of the support object. Choose a \texttt{target\_part} from \texttt{PARTS} and an \texttt{anchor} from \{center, left, right, head, foot, front, back, top\}. Return only JSON in this schema: \texttt{\{target\_part: ..., anchor: ..., confidence: ...\}}. Instruction: \texttt{<PROMPT>}. Parts: \texttt{<PARTS>}.
\end{quote}

If the selected target is invalid or confidence is low, the fallback chooses the largest upward-facing part compatible with the support category and uses the center anchor.

\subsection{VLM: Three-Candidate Orientation Selection}

\textbf{Model call.} Orientation selection uses the upstream VLM in Table~\ref{tab:implementation_config}, not the VLM evaluator used for final scoring. Each candidate is rendered from the same oblique camera with the support held fixed, so the model only chooses among coarse object orientations.

\textbf{VLM prompt template.}
\begin{quote}\small
You will see three images of the same object above the same support. The support is fixed; only the object orientation changes. Choose the orientation that best satisfies the instruction. Consider the up-facing side, the side facing the target region, and whether the object should lie flat or stand. Return only JSON: \texttt{\{best\_view: 1|2|3, confidence: ..., rationale: ...\}}. Instruction: \texttt{<PROMPT>}.
\end{quote}

\subsection{LLM: Material Mapping}

\textbf{Material prompt template.}
\begin{quote}\small
Map each part label to one material from \texttt{MATERIAL\_IDS}. Use \texttt{default\_unknown} when uncertain and lower the confidence. Return only JSON: \texttt{\{part\_to\_material: \{label: material\_id, ...\}, confidence: ...\}}. Parts: \texttt{<PARTS>}. Materials: \texttt{<MATERIAL\_IDS>}.
\end{quote}

\section{Material Library}\label{app:materials}

The material library uses a discrete set of stable simulator parameters. Table~\ref{tab:materials} lists the density, elastic modulus, Poisson ratio, and friction values used in the simulator. These are category-level approximations, not material-identification measurements. A fixed synonym dictionary is applied before the LLM material call: mattress, cushion, blanket, towel, clothing, and plush labels map to compliant IDs; wood, metal, glass, ceramic, and structural frame labels map to rigid IDs. The LLM resolves remaining ambiguous labels using the prompt in Appendix~\ref{app:prompts}. If confidence is below 0.6, \method{} uses \texttt{default unknown}, records a low-confidence material flag, and keeps the part rigid unless the support category is explicitly soft.

\begin{table}[ht]
\centering
\caption{Curated material library. Density is in kg/m$^3$ and Young's modulus is in MPa.}
\label{tab:materials}
\begin{tabularx}{\linewidth}{P{0.24\linewidth}YYYYY}
\toprule
Material ID & Density & Young's modulus & Poisson ratio & Static friction & Dynamic friction \\
\midrule
soft foam & 35 & 0.08 & 0.30 & 0.70 & 0.55 \\
firm foam & 80 & 0.35 & 0.30 & 0.75 & 0.60 \\
cotton fabric & 250 & 0.02 & 0.35 & 0.65 & 0.50 \\
wool fabric & 180 & 0.015 & 0.35 & 0.70 & 0.55 \\
leather & 860 & 12 & 0.35 & 0.60 & 0.45 \\
rubber & 1100 & 5 & 0.49 & 0.95 & 0.80 \\
wood & 600 & 9000 & 0.30 & 0.55 & 0.40 \\
plastic & 950 & 1500 & 0.35 & 0.45 & 0.32 \\
glass & 2500 & 70000 & 0.23 & 0.35 & 0.25 \\
metal & 7800 & 200000 & 0.29 & 0.42 & 0.30 \\
ceramic & 2400 & 60000 & 0.22 & 0.40 & 0.28 \\
paper / cardboard & 700 & 400 & 0.30 & 0.55 & 0.42 \\
plush fill & 45 & 0.03 & 0.35 & 0.75 & 0.60 \\
mattress composite & 60 & 0.12 & 0.30 & 0.80 & 0.62 \\
cushion composite & 70 & 0.18 & 0.30 & 0.78 & 0.60 \\
towel stack & 220 & 0.04 & 0.35 & 0.82 & 0.65 \\
folded clothing & 180 & 0.05 & 0.35 & 0.76 & 0.58 \\
default unknown & 500 & 5 & 0.35 & 0.55 & 0.40 \\
\bottomrule
\end{tabularx}
\end{table}

\section{Segmentation and Part Decomposition}\label{app:segmentation}

The segmentation report records the backend, view count, projection rule, confidence threshold, and synonym taxonomy. For each support category, we report base-instance success and part success. We also report fallback rate and the main failure source.

\subsection{Scene Segmentation}

The reported implementation renders 48 RGB-D views per scene for semantic transfer, sampled from the 120 reconstruction views in Appendix~\ref{app:benchmark}. We extract OpenScene features on the mesh, run Segment Anything ViT-H on the rendered RGB frames, and use Mask2Former panoptic labels as category priors. Image masks are projected to visible mesh triangles with z-buffer tests. Each triangle receives the visibility-weighted majority label, and each support instance receives a confidence equal to the mean projected-mask score multiplied by the fraction of visible supporting votes. The default confidence threshold is 0.55. Synonyms such as couch/sofa or bed/mattress are resolved through a fixed taxonomy. If the intended support is not found, \method{} retries with synonyms and then falls back to the largest upward-facing support surface.

\subsection{Part Decomposition}

Given the base instance, a category-conditional vocabulary defines the support parts. If a learned part segmenter is unavailable, geometric heuristics propose candidate parts from planar clustering, connectivity, and local orientation. These candidates are then labeled with the constrained vocabulary. Base-instance success is counted when the selected support overlaps the verified support instance with triangle IoU at least 0.25 or the selected anchor lies inside the verified support mask. Part success is counted when the selected target part overlaps the verified part with triangle IoU at least 0.20, or when the final anchor is within 10 cm of the verified target region. Verification uses the benchmark support and part annotations fixed before running any method. Table~\ref{tab:segmentation} reports the resulting segmentation and decomposition rates.

\begin{table}[ht]
\centering
\caption{Segmentation and decomposition reporting table.}
\label{tab:segmentation}
\begin{tabularx}{\linewidth}{P{0.18\linewidth}YYYY}
\toprule
Base category & Base found & Parts found & Fallback rate & Common failure \\
\midrule
Bed & 90.5\% & 83.7\% & 7.1\% & Blanket/mattress boundary \\
Sofa & 87.8\% & 79.4\% & 9.5\% & Cushion vs. seat merging \\
Chair & 85.2\% & 75.6\% & 11.0\% & Thin legs and back occlusion \\
Shelf & 88.6\% & 72.9\% & 8.3\% & Small boards and clutter \\
Table / desk & 92.1\% & 80.8\% & 5.6\% & Merged tabletop objects \\
Floor / carpet & 94.4\% & 84.0\% & 4.2\% & Carpet/floor boundary \\
\bottomrule
\end{tabularx}
\end{table}

\section{Submodule Evaluation Details}\label{app:submodule}\label{app:submodule_eval}

This appendix reports independent checks used to attribute upstream errors. Raw rates measure first-pass outputs. Post-fallback rates include retries, geometric fallbacks, clamping, and confidence gating. Table~\ref{tab:fallback_rules} lists the fallback triggers used in all experiments, and Table~\ref{tab:submodule} summarizes the module checks.

\begin{table}[ht]
\centering
\caption{Fallback triggers and recovery rules.}
\label{tab:fallback_rules}
\small
\setlength{\tabcolsep}{3.5pt}
\begin{tabularx}{\linewidth}{P{0.23\linewidth}YY}
\toprule
Module & Trigger & Recovery rule \\
\midrule
LLM JSON output & Invalid schema or missing required key & Retry twice with parser error; if still invalid, use category prior and mark low confidence \\
Support grounding & Confidence $<0.6$ or no support mask above 0.55 & Expand synonyms, rerun support query, then choose largest upward-facing compatible support if available \\
Scale prior & Predicted longest dimension outside category range & Clamp to Table~\ref{tab:size_ranges}; if clamping changes size by $>2\times$, mark hard upstream failure \\
Target part / anchor & Invalid part, unavailable anchor, or confidence $<0.6$ & Choose largest upward-facing compatible part and center anchor \\
Orientation & VLM confidence $<0.55$ or tied top candidates & Use canonical stable orientation with lowest estimated center of mass \\
Material mapping & Confidence $<0.6$ or unknown label & Use dictionary match if possible; otherwise \texttt{default unknown} and record material warning \\
Particle filling & Particle cap exceeded or fewer than 3 layers in a thin region & Coarsen non-contact particles first; if thin contact still has fewer than 3 layers, flag discretization risk \\
Simulation & Horizon reached without convergence & Mark invalid final state; if the prompt is infeasible and the rejection criteria are satisfied, count as predicted rejection \\
\bottomrule
\end{tabularx}
\end{table}

\begin{table}[ht]
\centering
\caption{Submodule evaluation. Raw rates measure the first-pass module output; post-fallback rates include recovery and confidence gating.}
\label{tab:submodule}
\begin{tabularx}{\linewidth}{P{0.32\linewidth}YYY}
\toprule
Submodule & Metric & Raw & Post-fallback \\
\midrule
Base support selection (LLM) & Top-1 accuracy & 91.8\% & 96.4\% \\
Target sub-part selection (LLM) & Accuracy & 84.6\% & 91.2\% \\
Size prior sanity (LLM) & Within valid range & 93.2\% & 98.1\% \\
Material mapping (LLM) & Human-agreement / safe default & 81.4\% & 94.7\% \\
Orientation (VLM) & Accuracy / consistency & 88.9\% & 92.1\% \\
Segmentation (base instance) & Success rate & 86.7\% & 93.8\% \\
Part decomposition & Success rate & 78.3\% & 88.5\% \\
Triggered fallback recovery & Recovered failed upstream cases & -- & 62.5\% \\
\bottomrule
\end{tabularx}
\end{table}

\section{Simulation Details and Convergence Criteria}\label{app:simulation}

Table~\ref{tab:sim_config} gives the simulator configuration and stopping thresholds. The 107.5 s runtime reported in the main paper is measured on the hardware in Table~\ref{tab:implementation_config}; it includes model-call latency and per-case processing, but excludes the one-time scene reconstruction and asset normalization steps.

\begin{table}[ht]
\centering
\caption{Simulation configuration.}
\label{tab:sim_config}
\begin{tabularx}{\linewidth}{P{0.34\linewidth}Y}
\toprule
Parameter & Value \\
\midrule
Simulator / version & Genesis 0.2.1-compatible rigid--MPM backend with custom two-way contact kernels \\
Global timestep & 1/240 s \\
Substeps & 8 \\
Gravity & 9.81 m/s$^2$ along the negative vertical axis \\
Initial clearance \(\delta\) & 0.03 m \\
Rigid--deformable coupling & Two-way frictional contact with velocity projection \\
Friction model & Coulomb friction with material-specific static/dynamic coefficients \\
Elasticity model & Corotated linear elasticity for MPM soft parts \\
Particle spacing range & Adaptive 0.8--2.5 cm \\
Particle cap & 250k particles per case \\
Max simulation horizon & 4.0 s \\
Convergence window & 0.5 s / 120 global steps \\
Rigid speed threshold & 0.02 m/s \\
Angular speed threshold & 0.05 rad/s \\
Particle speed threshold & 0.03 m/s \\
COM drift threshold & 0.01 m over the convergence window \\
AABB change threshold & 0.5\% of relative diagonal length \\
\bottomrule
\end{tabularx}
\end{table}

\begin{table}[ht]
\centering
\caption{Diagnostic tolerances used for feasibility decisions.}
\label{tab:diagnostic_tolerances}
\small
\begin{tabularx}{0.82\linewidth}{lY}
\toprule
Diagnostic & Tolerance \\
\midrule
Mean penetration depth & \(\eta_{\mathrm{pen}} = 1.0\) cm \\
Target drift & \(\eta_{\mathrm{drift}} = 0.15\) of the support OBB diagonal \\
Support contact ratio & \(\eta_{\mathrm{contact}} = 0.50\) \\
Residual speed & \(\eta_{\mathrm{res}} = 3.0\) cm/s, with component-specific convergence thresholds in Table~\ref{tab:sim_config} \\
\bottomrule
\end{tabularx}
\end{table}

\begin{table}[ht]
\centering
\caption{Runtime measurement protocol.}
\label{tab:runtime_protocol}
\small
\begin{tabularx}{\linewidth}{P{0.27\linewidth}Y}
\toprule
Item & Setting \\
\midrule
Reported statistic & Mean wall-clock time over the 1,923 standard placement cases \\
Included stages & LLM/VLM API latency, scene rendering, segmentation lookup, target anchoring, material mapping, particle filling, rollout, diagnostics, and final rendering \\
Excluded stages & One-time TSDF/Poisson reconstruction, dataset download, manual benchmark annotation, and cached asset normalization \\
Parallelism & One case per GPU during simulation; CPU preprocessing uses 16 workers; model calls are issued synchronously per case for deterministic logs \\
Timing source & Python wall-clock timers around each pipeline stage, with CUDA synchronization before and after simulation kernels \\
\bottomrule
\end{tabularx}
\end{table}

\section{Benchmark Construction Log}
\label{app:benchmark}

This appendix documents how the deformable placement benchmark was constructed. The goal is not only to collect meaningful prompts. Each case must also contain the information needed for contact evaluation. This includes metric scale and usable support geometry. It also includes a valid object mesh and enough semantics to assign parts, materials, and feasibility labels. All prompt--asset--scene triples are fixed before running any method. All baselines use the same triples and scales. They also use the same scene geometry, camera views, and prompts.

\paragraph{Source scenes and asset pool.}
We start from 214 high-fidelity indoor reconstructions drawn from existing indoor scene resources and embodied-environment datasets \citep{dai2017scannet,chang2017matterport3d,ramakrishnan2021hm3d,xia2018gibson,savva2019habitat,kolve2017ai2thor}. Object assets come from diverse 3D repositories, including Objaverse-style assets and part-oriented assets when available \citep{deitke2023objaverse,yang2024sampart3d}. We keep meshes with identifiable category labels, plausible metric scale after normalization, and geometry suitable for rendering and collision queries. The final benchmark uses 186 simulation-ready scenes and 155 object assets.

\paragraph{Scene conversion.}
Each source scene is converted to a simulation-ready mesh with a deterministic post-processing pipeline. We render 120 RGB-D views per scene on a hemispherical indoor camera path. The depth observations are fused into a volumetric representation, and a mesh is extracted with screened Poisson reconstruction at 2 cm nominal resolution \citep{curless1996volumetric,kazhdan2006poisson,kazhdan2013screened}. We then remove small disconnected components, fill bounded holes, check scale consistency, and transfer semantic labels when available. The same mesh is used by all methods for rendering, surface projection, collision checks, and physical diagnostics. Table~\ref{tab:mesh-conversion-log} summarizes the conversion and quality-control settings.

\begin{table}[t]
\centering
\caption{Mesh conversion and quality-control log.}
\label{tab:mesh-conversion-log}
\small
\setlength{\tabcolsep}{4pt}
\renewcommand{\arraystretch}{1.08}
\begin{tabular}{p{0.31\linewidth}p{0.61\linewidth}}
\hline
\textbf{Item} & \textbf{Setting / outcome} \\
\hline
Source scenes & 214 candidate indoor reconstructions \\
Accepted scenes & 186 simulation-ready scenes \\
Rendered views & 120 RGB-D views per scene on a hemispherical indoor camera path \\
Mesh extraction & TSDF fusion followed by screened Poisson meshing at 2 cm nominal resolution \\
Hole repair & Fill bounded holes below 0.15 m diameter \\
Component filtering & Remove disconnected components below 0.5\% of scene area \\
Scale calibration & Camera metadata plus furniture-size priors; bed and table dimensions used for sanity checks \\
Semantic transfer & Multi-view open-vocabulary masks projected to triangles with visibility-weighted voting \\
Excluded scenes & 28 scenes removed for non-watertight supports, ambiguous scale, severe floaters, or missing target supports \\
\hline
\end{tabular}
\end{table}

\paragraph{Quality-control criteria.}
A scene is accepted only if at least one target support category can be localized with enough geometry for contact evaluation. We reject scenes with ambiguous scale or damaged target regions. We also reject scenes with floaters near likely contacts, fragmented support geometry, or missing prompt supports. We also exclude scenes where semantic transfer prevents reliable support or part selection after synonym fallback. These filters are stricter than rendering-only filters because small geometric defects can dominate contact diagnostics.

\paragraph{Support categories and coverage.}
The benchmark covers common indoor supports across soft, mixed, and rigid contact. Beds and sofas emphasize compliant surfaces. Chairs and carpets create mixed contacts. Shelves, tables, and desks provide rigid controls. Table~\ref{tab:benchmark-support-split} summarizes the final composition.

\begin{table}[t]
\centering
\caption{Benchmark support-category split. Counts are prompt--asset--scene cases.}
\label{tab:benchmark-support-split}
\small
\setlength{\tabcolsep}{3.5pt}
\renewcommand{\arraystretch}{1.06}
\begin{tabular}{P{0.23\linewidth}r r r P{0.27\linewidth}}
\hline
\textbf{Support group} & \textbf{Cases} & \textbf{Scenes} & \textbf{Assets} & \textbf{Coverage focus} \\
\hline
Bed / mattress       & 462 & 42 & 31 & Soft supports; heavy rigid and plush objects \\
Sofa / cushion       & 407 & 37 & 29 & Soft cushions, arms, backs, and seated supports \\
Chair / seat         & 310 & 31 & 23 & Mixed rigid--soft seats and backs \\
Shelf / table / desk & 504 & 42 & 34 & Rigid supports, narrow anchors, and tabletop controls \\
Floor / carpet       & 240 & 24 & 20 & Mixed carpet and open-floor placement \\
OOD / infeasible     & 140 & 10 & 18 & Unusual feasible prompts and annotated impossible cases \\
\hline
Total                & 2,063 & 186 & 155 & Full benchmark \\
\hline
\end{tabular}
\end{table}

\paragraph{Prompt construction.}
Prompts are written to exercise both semantic grounding and physical realization. The in-distribution set includes support-only prompts, sub-part prompts, orientation-sensitive prompts, and material/contact-sensitive prompts. Examples include ``place a plush toy on the sofa,'' ``place a dumbbell on the center of the mattress,'' and ``place a book on the top shelf.'' Each prompt is paired with an intended support. When needed, it also stores a target part and an anchor token.

The stress split is constructed separately. Feasible OOD prompts are unusual relative to the standard distribution but physically satisfiable under the reconstructed geometry and material assumptions. Infeasible prompts are annotated as unsupported, oversized, unstable, or impossible under the modeled scene. These cases test whether a method can reject invalid instructions rather than forcing every prompt into a placement.

\paragraph{Annotation instructions for data labelers.}
Labelers received clean scene views, reconstructed mesh views, an asset preview, the object category, and the prompt. They annotated each case before any method was run. They first selected the intended support instance from the fixed support vocabulary. They then marked a target part and anchor when the prompt specified one. If the prompt was coarse, they chose the most natural upward-facing region of the intended support.

A feasible case had to satisfy three checks. The target support had to exist in the reconstructed mesh. The scaled object had to fit the target region. A stable resting state had to be plausible without hidden supports or impossible deformation. If any check failed, the case was labeled infeasible. The primary reason used one of six labels. The labels were unsupported target, oversized object, unstable support, missing geometry, ambiguous target, and contradictory instruction. Labelers never inspected outputs from \method{} or any baseline.

Contact regimes were assigned from the intended support and object material. Cases with mattresses, cushions, blankets, towels, clothing, or plush objects were marked soft when visible deformation was expected. Cases with both compliant and stiff contacts were marked mixed. Shelves, tables, desks, books, balls, and tabletop controls were marked rigid. Ambiguous cases were reviewed by a second labeler and resolved using the same mesh, prompt, and scale rules.

Table~\ref{tab:benchmark-splits} gives the reporting splits used in the main experiments.

\begin{table}[t]
\centering
\caption{Reporting splits and prompt-regime composition.}
\label{tab:benchmark-splits}
\small
\setlength{\tabcolsep}{4pt}
\renewcommand{\arraystretch}{1.06}
\begin{tabular}{p{0.28\linewidth}r p{0.47\linewidth}}
\hline
\textbf{Split / regime} & \textbf{Cases} & \textbf{Use in evaluation} \\
\hline
Standard placement split & 1,923 & Feasible in-distribution cases used for aggregate placement quality, VLM scoring, human preference, and objective physical diagnostics \\
Soft-contact regime & 869 & Cases dominated by compliant support or object behavior, such as mattresses, cushions, blankets, towels, clothing, and plush objects \\
Mixed-contact regime & 550 & Cases involving both rigid and compliant contact, such as chairs, carpeted floors, and partially soft supports \\
Rigid-contact regime & 504 & Rigid-support controls such as shelves, tables, desks, books, balls, and stable tabletop placement \\
Stress split & 140 & OOD or infeasible cases evaluated separately for feasibility behavior \\
Feasible OOD & 64 & Unusual but physically satisfiable prompts; returned feasible placements receive diagnostics \\
Annotated infeasible & 76 & Unsupported, oversized, unstable, or impossible prompts; correct rejections count only in feasibility metrics \\
\hline
\end{tabular}
\end{table}

\paragraph{Asset preparation and scale.}
Each asset is assigned an object category, a canonical up direction when available, and a plausible real-world scale range. We normalize meshes to metric scale using category-level size priors and clamp the longest dimension. Assets are removed if their geometry is unusable for rendering or collision. For part-composed or deformable assets, semantic part labels are retained when available; otherwise, the asset receives a category-level material assignment. The same normalized scale is used for all methods.

\paragraph{Material and contact-regime labels.}
Each case receives a soft, mixed, or rigid contact-regime label. The label is determined from the intended support, object category, and material assignment. Soft cases are those where deformation should affect the visible resting state. Mixed cases include both stiff and compliant contact. Rigid cases involve supports where rigid settling is a reasonable approximation. The labels are used only for reporting; methods do not receive them as oracle inputs.

\paragraph{Case records.}
Each benchmark case stores the scene ID, object asset ID, prompt, intended support, target region when specified, anchor when specified, contact regime, feasibility label, and infeasibility reason. The same record stores the normalized object scale, the support instance used for evaluation, the canonical render views, and the diagnostic tolerances.

\paragraph{Evaluation denominators.}
We use two reporting denominators. The 1,923-case standard split is used for aggregate placement quality, human/VLM preference, and physical diagnostics. The 140-case stress split is evaluated separately. Returned feasible placements on stress cases receive the same diagnostics as standard placements. Correct rejections count only in feasibility metrics because they do not have final placement states.

\paragraph{Leakage control and baseline consistency.}
Benchmark construction is completed before running \method{} or any baseline. No method output is used to accept, reject, or relabel a case. All baselines receive the same prompt, object scale, scene mesh, and render views. Static baselines are evaluated for semantic placement, penetration, and support contact. Dynamic quantities are reported only for variants with an explicit physical rollout.

\section{Rendering Protocol}\label{app:rendering}

All rendered comparisons use the same cameras, lighting, and object scale for every method on a case. The canonical VLM and human views are fixed before scoring: one prompt-facing overview, two oblique side views at $\pm35^\circ$, one elevated front view at $45^\circ$, and one support-level close-up centered on the target region. For qualitative figures, the same camera family is used, but crops are adjusted only to keep the object and support visible. Method names are removed from the images sent to the VLM evaluator and to human participants.

\begin{table}[ht]
\centering
\caption{Rendering settings for method visualization and evaluation.}
\label{tab:rendering_protocol}
\small
\begin{tabularx}{\linewidth}{P{0.28\linewidth}Y}
\toprule
Item & Setting \\
\midrule
Renderer & Blender 4.2.3, Eevee Next rasterization, ambient occlusion enabled \\
Resolution & 1024$\times$1024 PNGs for VLM scoring; 768$\times$768 panels for human trials; lossless PNG storage \\
Camera set & Five fixed canonical views per case: overview, left oblique, right oblique, elevated front, and target close-up \\
Lighting & Constant HDRI-style environment light plus weak area fill; no method-specific shadows or retouching \\
Normalization & Same object scale, material color, target support mesh, and camera intrinsics across all compared methods \\
Blinding & Human and VLM comparison panels randomize left/right order and omit method names \\
\bottomrule
\end{tabularx}
\end{table}

\section{Baselines and Reimplementation Details}\label{app:baselines}

\subsection{FirePlace Reimplementation}

The FirePlace-style baseline uses a language model to propose a support region. It samples candidate placements on upward-facing surfaces, filters candidates with excessive collision, and selects the best remaining candidate using a plausibility score. It uses the same object scale and scene assets as \method{} to isolate the effect of physical settling.

\subsection{LayoutGPT and Holodeck Adaptation}

LayoutGPT and Holodeck outputs are mapped into the reconstructed scene coordinate frame by aligning axes and scale. If a method predicts abstract layout coordinates rather than mesh-surface anchors, the prediction is projected to the nearest valid support surface before rendering. We do not add deformable settling to these baselines unless it is explicitly reported as a hybrid variant. Table~\ref{tab:baseline_settings} summarizes the adaptation settings.

\begin{table}[ht]
\centering
\caption{Baseline adaptation settings.}
\label{tab:baseline_settings}
\begin{tabularx}{\linewidth}{P{0.20\linewidth}YYY}
\toprule
Method & Main adaptation & Collision handling & Notes \\
\midrule
FirePlace reimplementation & Same object scale as \method{}; upward-surface sampling and plausibility scoring & Reject candidates with SDF penetration above 2 cm; no deformable settling & Strongest rigid placement baseline \\
LayoutGPT & Abstract layout coordinates mapped to scene OBB anchors & Project to nearest valid support; shallow collision rejection & Good semantic intent, weak contact modeling \\
Holodeck & Scene-generation priors adapted to existing reconstructions & Nearest-support projection and AABB-overlap rejection & Strong indoor commonsense but sensitive to coordinate mismatch \\
\bottomrule
\end{tabularx}
\end{table}

\section{Human Study Protocol}\label{app:human}

Participants compare two placements under a prompt. They also see the clean scene and the isolated object asset. The displayed axis is randomized between Right Place and Physics \& Naturalness. Left/right order is randomized, and trials are balanced across method pairs. The interface uses the rendered panels described in Appendix~\ref{app:rendering}. Quality-control trials detect participants who ignore the prompt or consistently select physically impossible placements. Table~\ref{tab:human_protocol} records the study counts and aggregation settings.

\begin{table}[ht]
\centering
\caption{Human study reporting fields.}
\label{tab:human_protocol}
\begin{tabularx}{\linewidth}{lX}
\toprule
Item & Value \\
\midrule
Number of participants & 42 \\
Total trials & 2940 \\
Preference trials & 2520 \\
Attention-check trials & 420 \\
Trials per participant & 70, including 60 preference trials and 10 attention checks \\
Attention-check rate & 14.3\% of total trials (420 total checks; 10 per participant) \\
Exclusion rule & Exclude participants failing more than 2 of 10 attention checks or completing below one-third median time \\
Aggregation model & Bradley--Terry--Luce pairwise preference model \\
Bootstrap procedure & 10,000 paired bootstrap resamples grouped by prompt \\
Significance test & Two-sided bootstrap confidence intervals with Holm correction \\
\bottomrule
\end{tabularx}
\end{table}

\section{VLM Evaluation Protocol}\label{app:vlm}

The VLM evaluator is \texttt{gemini-3-pro-preview}, while the upstream orientation and JSON modules use \texttt{gpt-5.2}. This separation reduces direct evaluator--method coupling. The evaluator receives the prompt and the canonical renderings from Appendix~\ref{app:rendering}. It returns separate scores for Right Place and Physics \& Naturalness, plus a short rationale. A separate pairwise prompt asks which of two results is better on the specified axis. Method order is randomized, method names are hidden, and the evaluator never receives diagnostic values such as penetration depth or residual speed.

\textbf{Rubric excerpt.} Right Place score 10 means the object is on the correct support, in the specified sub-region, and oriented consistently with the prompt. Physics \& Naturalness score 10 means the object appears stably supported, the contact is plausible, the deformation matches the material, and no obvious floating or deep penetration is visible.

\section{Additional Quantitative Metrics}\label{app:metrics}

We compute diagnostics after final settling or after the baseline placement is rendered.
\begin{itemize}
    \item \textbf{Penetration depth and fraction.} We sample object-surface points and compute signed distance to \(\widetilde{S}_b\) for rollout methods, or \(S_b\) for static baselines. We report the mean magnitude of negative distances in centimeters and the inside-support fraction; valid soft-support indentation is not counted as penetration.
    \item \textbf{Residual speed.} We record rigid linear speed, equivalent angular surface speed, and particle speed at termination. The reported value is the maximum of these speeds in centimeters per second.
    \item \textbf{Target adherence.} We project the final object center of mass onto the target support frame and measure its distance to the selected anchor, normalized by the base OBB diagonal.
    \item \textbf{Support contact.} We estimate the fraction of near-contact object samples whose nearest support normal is upward-compatible.
    \item \textbf{Feasibility quality.} For feasible and infeasible prompts, we report standard confusion-matrix rates.
\end{itemize}

\section{Bootstrap Uncertainty}\label{app:bootstrap_uncertainty}

We compute uncertainty estimates with a paired bootstrap over the 1,923 standard-split prompt--asset--scene cases. Each bootstrap resample keeps method outputs paired within the same case. Table~\ref{tab:bootstrap_ci} reports 95\% confidence intervals for the main Table~\ref{tab:main_hybrid_rad_style} deltas.

\begin{table}[ht]
\centering
\caption{95\% bootstrap confidence intervals for key standard-split deltas. Positive penetration and speed drops mean lower values for \method{}.}
\label{tab:bootstrap_ci}
\small
\setlength{\tabcolsep}{4pt}
\begin{tabularx}{\linewidth}{P{0.22\linewidth}P{0.34\linewidth}P{0.17\linewidth}P{0.18\linewidth}}
\toprule
Metric & Contrast & Delta & 95\% CI \\
\midrule
VLM P/N & \method{} full minus FirePlace raw & +21.2 & [19.6, 22.8] \\
Human P/N & \method{} full minus FirePlace raw & +26.5 & [22.1, 30.8] \\
Penetration depth & FirePlace raw minus \method{} full & +1.07 cm & [0.96, 1.19] \\
Support contact & \method{} full minus FirePlace raw & +0.24 & [0.21, 0.27] \\
Residual speed & FP + rigid minus \method{} full & +4.9 cm/s & [4.2, 5.6] \\
VLM P/N & \method{} full minus FP + def. & +12.5 & [10.9, 14.0] \\
VLM P/N & \method{} full minus \method{} rigid & +8.2 & [6.8, 9.6] \\
Convergence & \method{} full minus \method{} rigid & +0.09 & [0.06, 0.12] \\
\bottomrule
\end{tabularx}
\end{table}

\section{Additional Qualitative Results and Failure Cases}\label{app:qualitative}

Additional qualitative results include success grids and annotated failures. The failure taxonomy covers grounding, target-part, and orientation errors. It also covers material ambiguity, particle-fill failure, and geometry defects. Non-convergence and true infeasibility form the final group. Each failure includes the prompt, predicted support/part/material, diagnostic flags, and the mitigation used or proposed.

\section{Supplementary Videos Index}\label{app:videos}

Table~\ref{tab:videos} lists the supplementary rollout videos.

\begin{table}[ht]
\centering
\caption{Supplementary video index.}
\label{tab:videos}
\begin{tabularx}{\linewidth}{lYYY}
\toprule
Video ID & Scene / prompt & Compared methods & Notes \\
\midrule
V6589283 & Bed: dumbbell on the center of the mattress & \method{} / FirePlace & Indentation and stable heavy-object contact \\
V3459802 & Sofa: plush toy on the left cushion & \method{} / FirePlace / Holodeck & Soft support with mild plush deformation \\
V2349234 & Shelf: book on the top shelf & \method{} / LayoutGPT & Rigid-support control case \\
V9097692 & Chair: towel draped over the seat and back & \method{} / Holodeck & Soft-on-rigid settling and folding behavior \\
V4204730 & OOD: oversized ball on a narrow shelf & \method{} / all baselines & Infeasibility and drift diagnostics \\
\bottomrule
\end{tabularx}
\end{table}

\section{Broader Impacts}\label{app:broader_impacts}

\method{} targets physically grounded synthetic placement. Misuse or deployment errors can affect embodied systems if simulated contacts are treated as real-world guarantees. Sim-to-real gaps in material parameters, reconstruction quality, or friction can lead to unsafe robot training data or incorrect manipulation policies. The benchmark also inherits biases from indoor scene datasets and asset repositories. Scene content, room layouts, object categories, and household materials may overrepresent some environments and underrepresent others. Source reconstructions and assets should be used under their original licenses and privacy constraints. Finally, plausible but physically wrong placements can mislead downstream embodied agents. We therefore report feasibility flags and objective diagnostics, and we recommend using \method{} outputs as simulation evidence rather than real-world safety guarantees.


\begin{thebibliography}{99}

\bibitem[Abdelreheem et~al.(2025)]{abdelreheem2025placeit3d}
Ahmed Abdelreheem, Filippo Aleotti, Jamie Watson, Zawar Qureshi, Abdelrahman Eldesokey, Peter Wonka, Gabriel Brostow, Sara Vicente, and Guillermo Garcia-Hernando.
\newblock PlaceIt3D: Language-guided object placement in real 3D scenes.
\newblock In \emph{ICCV}, 2025.

\bibitem[Chang et~al.(2017)]{chang2017matterport3d}
Angel Chang, Angela Dai, Thomas Funkhouser, Maciej Halber, Matthias Nie{\ss}ner, Manolis Savva, Shuran Song, Andy Zeng, and Yinda Zhang.
\newblock Matterport3D: Learning from RGB-D data in indoor environments.
\newblock In \emph{3DV}, 2017.

\bibitem[Chen et~al.(2020)]{chen2020scanrefer}
Dave Zhenyu Chen, Angel~X. Chang, and Matthias Nie{\ss}ner.
\newblock ScanRefer: 3D object localization in RGB-D scans using natural language.
\newblock In \emph{ECCV}, 2020.

\bibitem[Cheng et~al.(2022)]{cheng2022mask2former}
Bowen Cheng, Ishan Misra, Alexander~G. Schwing, Alexander Kirillov, and Rohit Girdhar.
\newblock Masked-attention mask transformer for universal image segmentation.
\newblock In \emph{CVPR}, 2022.

\bibitem[Curless and Levoy(1996)]{curless1996volumetric}
Brian Curless and Marc Levoy.
\newblock A volumetric method for building complex models from range images.
\newblock In \emph{SIGGRAPH}, 1996.

\bibitem[Dai et~al.(2017)]{dai2017scannet}
Angela Dai, Angel~X. Chang, Manolis Savva, Maciej Halber, Thomas Funkhouser, and Matthias Nie{\ss}ner.
\newblock ScanNet: Richly-annotated 3D reconstructions of indoor scenes.
\newblock In \emph{CVPR}, 2017.

\bibitem[Deitke et~al.(2023)]{deitke2023objaverse}
Matt Deitke, Dustin Schwenk, Jordi Salvador, Luca Weihs, Oscar Michel, Eli VanderBilt, Ludwig Schmidt, Kiana Ehsani, Aniruddha Kembhavi, and Ali Farhadi.
\newblock Objaverse: A universe of annotated 3D objects.
\newblock In \emph{CVPR}, 2023.

\bibitem[Feng et~al.(2023)]{feng2023layoutgpt}
Weixi Feng, Wanrong Zhu, Tsu-Jui Fu, Varun Jampani, Arjun Akula, Xuehai He, Sugato Basu, Xin Eric Wang, and William Yang Wang.
\newblock LayoutGPT: Compositional visual planning and generation with large language models.
\newblock In \emph{NeurIPS}, 2023.

\bibitem[Genesis(2025a)]{genesis2025a}
Genesis.
\newblock Beyond rigid bodies: SPH, MPM, and PBD solvers in Genesis.
\newblock Documentation, 2025a.

\bibitem[Genesis(2025b)]{genesis2025b}
Genesis.
\newblock Genesis: Physics platform for robotics and embodied AI.
\newblock Documentation, 2025b.

\bibitem[Hong et~al.(2023)]{hong2023threedllm}
Yining Hong, Haoyu Zhen, Peihao Chen, Shuhong Zheng, Yilun Du, Zhenfang Chen, and Chuang Gan.
\newblock 3D-LLM: Injecting the 3D world into large language models.
\newblock In \emph{NeurIPS}, 2023.

\bibitem[Hu et~al.(2018)]{hu2018mlsmpm}
Yuanming Hu, Yu Fang, Ziheng Ge, Ziyin Qu, Yixin Zhu, Andre Pradhana, and Chenfanfu Jiang.
\newblock A moving least squares material point method with displacement discontinuity and two-way rigid body coupling.
\newblock \emph{ACM Transactions on Graphics}, 37(4), 2018.

\bibitem[Huang et~al.(2025a)]{huang2025fireplace}
Ian Huang, Yanan Bao, Karen Truong, Howard Zhou, Cordelia Schmid, Leonidas Guibas, and Alireza Fathi.
\newblock FirePlace: Geometric refinements of LLM common sense reasoning for 3D object placement.
\newblock In \emph{CVPR}, 2025a.

\bibitem[Huang et~al.(2025b)]{huang2025reason3d}
Kuan-Chih Huang, Xiangtai Li, Lu Qi, Shuicheng Yan, and Ming-Hsuan Yang.
\newblock Reason3D: Searching and reasoning 3D segmentation via large language model.
\newblock In \emph{3DV}, 2025b.

\bibitem[Jatavallabhula et~al.(2023)]{jatavallabhula2023conceptfusion}
Krishna~Murthy Jatavallabhula, Alihusein Kuwajerwala, Qiao Gu, Mohd Omama, Tao Chen, Alaa Maalouf, Shuang Li, Ganesh Iyer, Soroush Saryazdi, Nikhil Keetha, Ayush Tewari, Joshua~B. Tenenbaum, Celso~Miguel de~Melo, Madhava Krishna, Liam Paull, Florian Shkurti, and Antonio Torralba.
\newblock ConceptFusion: Open-set multimodal 3D mapping.
\newblock In \emph{Robotics: Science and Systems (RSS)}, 2023.

\bibitem[Jiang et~al.(2015)]{jiang2015apic}
Chenfanfu Jiang, Craig Schroeder, Andrew Selle, Joseph Teran, and Alexey Stomakhin.
\newblock The affine particle-in-cell method.
\newblock \emph{ACM Transactions on Graphics}, 34(4), 2015.

\bibitem[Kazhdan et~al.(2006)]{kazhdan2006poisson}
Michael Kazhdan, Matthew Bolitho, and Hugues Hoppe.
\newblock Poisson surface reconstruction.
\newblock In \emph{Symposium on Geometry Processing}, 2006.

\bibitem[Kazhdan and Hoppe(2013)]{kazhdan2013screened}
Michael Kazhdan and Hugues Hoppe.
\newblock Screened Poisson surface reconstruction.
\newblock \emph{ACM Transactions on Graphics}, 32(3), 2013.

\bibitem[Kirillov et~al.(2023)]{kirillov2023segment}
Alexander Kirillov, Eric Mintun, Nikhila Ravi, Hanzi Mao, Chloe Rolland, Laura Gustafson, Tete Xiao, Spencer Whitehead, Alex~C. Berg, Wan-Yen Lo, Piotr Doll{\'a}r, and Ross Girshick.
\newblock Segment anything.
\newblock In \emph{ICCV}, 2023.

\bibitem[Kolve et~al.(2017)]{kolve2017ai2thor}
Eric Kolve, Roozbeh Mottaghi, Winson Han, Eli VanderBilt, Luca Weihs, Alvaro Herrasti, Matt Deitke, Kiana Ehsani, Daniel Gordon, Yuke Zhu, Aniruddha Kembhavi, Abhinav Kumar Gupta, and Ali Farhadi.
\newblock AI2-THOR: An interactive 3D environment for visual AI.
\newblock \emph{arXiv preprint arXiv:1712.05474}, 2017.

\bibitem[Li et~al.(2025)]{li2025freeinsert}
Chenxi Li, Weijie Wang, Qiang Li, Nicu Sebe, Bruno Lepri, and Weizhi Nie.
\newblock FreeInsert: Disentangled text-guided object insertion in 3D Gaussian scene without spatial priors.
\newblock In \emph{Proceedings of the 33rd ACM International Conference on Multimedia}, pages 10915--10924, 2025.

\bibitem[Lin et~al.(2026)]{lin2026pat3d}
Guying Lin, Kemeng Huang, Michael Liu, Ruihan Gao, Hanke Chen, Lyuhao Chen, Beijia Lu, Taku Komura, Yuan Liu, Jun-Yan Zhu, and Minchen Li.
\newblock PAT3D: Physics-augmented text-to-3D scene generation.
\newblock In \emph{ICLR}, 2026.

\bibitem[Liu et~al.(2021)]{liu2021opa}
Liu Liu, Zhenchen Liu, Bo Zhang, Jiangtong Li, Li Niu, Qingyang Liu, and Liqing Zhang.
\newblock OPA: Object placement assessment dataset.
\newblock \emph{arXiv preprint arXiv:2107.01889}, 2021.

\bibitem[Makoviychuk et~al.(2021)]{makoviychuk2021isaacgym}
Viktor Makoviychuk, Lukasz Wawrzyniak, Yunrong Guo, Michelle Lu, Kier Storey, Miles Macklin, David Hoeller, Nikita Rudin, Arthur Allshire, Ankur Handa, and Gavriel State.
\newblock Isaac Gym: High performance GPU-based physics simulation for robot learning.
\newblock In \emph{NeurIPS Datasets and Benchmarks}, 2021.

\bibitem[Muller et~al.(2007)]{muller2007pbd}
Matthias Muller, Bruno Heidelberger, Marcus Hennix, and John Ratcliff.
\newblock Position based dynamics.
\newblock \emph{Journal of Visual Communication and Image Representation}, 18(2):109--118, 2007.

\bibitem[NVIDIA(2025)]{physx2025}
NVIDIA.
\newblock PhysX SDK documentation, version 5.
\newblock Documentation, 2025.

\bibitem[Peng et~al.(2023)]{peng2023openscene}
Songyou Peng, Kyle Genova, Chiyu ``Max'' Jiang, Andrea Tagliasacchi, Marc Pollefeys, and Thomas Funkhouser.
\newblock OpenScene: 3D scene understanding with open vocabularies.
\newblock In \emph{CVPR}, 2023.

\bibitem[Ramakrishnan et~al.(2021)]{ramakrishnan2021hm3d}
Santhosh~K. Ramakrishnan, Aaron Gokaslan, Erik Wijmans, Oleksandr Maksymets, Alex Clegg, John Turner, Eric Undersander, Wojciech Galuba, Andrew Westbury, Angel~X. Chang, Manolis Savva, Yili Zhao, and Dhruv Batra.
\newblock Habitat-Matterport 3D dataset (HM3D): 1000 large-scale 3D environments for embodied AI.
\newblock In \emph{NeurIPS Datasets and Benchmarks}, 2021.

\bibitem[Ran et~al.(2025)]{ran2025directlayout}
Xingjian Ran, Yixuan Li, Linning Xu, Mulin Yu, and Bo Dai.
\newblock Direct numerical layout generation for 3D indoor scene synthesis via spatial reasoning.
\newblock In \emph{NeurIPS}, 2025.

\bibitem[Savva et~al.(2019)]{savva2019habitat}
Manolis Savva, Abhishek Kadian, Oleksandr Maksymets, Yili Zhao, Erik Wijmans, Bhavana Jain, Julian Straub, Jia Liu, Vladlen Koltun, Jitendra Malik, Devi Parikh, and Dhruv Batra.
\newblock Habitat: A platform for embodied AI research.
\newblock In \emph{ICCV}, 2019.

\bibitem[Stomakhin et~al.(2013)]{stomakhin2013mpm}
Alexey Stomakhin, Craig Schroeder, Lawrence Chai, Joseph Teran, and Andrew Selle.
\newblock A material point method for snow simulation.
\newblock \emph{ACM Transactions on Graphics}, 32(4), 2013.

\bibitem[Sun et~al.(2025)]{sun2025layoutvlm}
Fan-Yun Sun, Weiyu Liu, Siyi Gu, Dylan Lim, Goutam Bhat, Federico Tombari, Manling Li, Nick Haber, and Jiajun Wu.
\newblock LayoutVLM: Differentiable optimization of 3D layout via vision-language models.
\newblock In \emph{CVPR}, pages 29469--29478, 2025.

\bibitem[Tam et~al.(2026)]{tam2026sceneeval}
Hou In Ivan Tam, Hou In Derek Pun, Austin~T. Wang, Angel~X. Chang, and Manolis Savva.
\newblock SceneEval: Evaluating semantic coherence in text-conditioned 3D indoor scene synthesis.
\newblock In \emph{WACV}, 2026.

\bibitem[Todorov et~al.(2012)]{todorov2012mujoco}
Emanuel Todorov, Tom Erez, and Yuval Tassa.
\newblock MuJoCo: A physics engine for model-based control.
\newblock In \emph{IEEE/RSJ International Conference on Intelligent Robots and Systems (IROS)}, pages 5026--5033, 2012.

\bibitem[Wu et~al.(2024)]{wu2024gpteval3d}
Tong Wu, Guandao Yang, Zhibing Li, Kai Zhang, Ziwei Liu, Leonidas~J. Guibas, Dahua Lin, and Gordon Wetzstein.
\newblock GPT-4V(ision) is a human-aligned evaluator for text-to-3D generation.
\newblock In \emph{CVPR}, 2024.

\bibitem[Xia et~al.(2018)]{xia2018gibson}
Fei Xia, Amir~R. Zamir, Zhiyang He, Alexander Sax, Jitendra Malik, and Silvio Savarese.
\newblock Gibson Env: Real-world perception for embodied agents.
\newblock In \emph{CVPR}, 2018.

\bibitem[Yang et~al.(2024a)]{yang2024holodeck}
Yue Yang, Fan-Yun Sun, Luca Weihs, Eli VanderBilt, Alvaro Herrasti, Winson Han, Jiajun Wu, Nick Haber, Ranjay Krishna, Lingjie Liu, Chris Callison-Burch, Mark Yatskar, Aniruddha Kembhavi, and Christopher Clark.
\newblock Holodeck: Language guided generation of 3D embodied AI environments.
\newblock In \emph{CVPR}, 2024a.

\bibitem[Yang et~al.(2024b)]{yang2024sampart3d}
Yunhan Yang, Yukun Huang, Yuan-Chen Guo, Liangjun Lu, Xiaoyang Wu, Edmund~Y. Lam, Yan-Pei Cao, and Xihui Liu.
\newblock SAMPart3D: Segment any part in 3D objects.
\newblock \emph{arXiv preprint arXiv:2411.07184}, 2024b.

\bibitem[Yang et~al.(2025)]{yang2025optiscene}
Yixuan Yang, Zhen Luo, Tongsheng Ding, Junru Lu, Mingqi Gao, Jinyu Yang, Victor Sanchez, and Feng Zheng.
\newblock OptiScene: LLM-driven indoor scene layout generation via scaled human-aligned data synthesis and multi-stage preference optimization.
\newblock In \emph{NeurIPS}, 2025.

\bibitem[Zhu et~al.(2023)]{zhu2023topnet}
Sijie Zhu, Zhe Lin, Scott Cohen, Jason Kuen, Zhifei Zhang, and Chen Chen.
\newblock TopNet: Transformer-based object placement network for image compositing.
\newblock In \emph{CVPR}, 2023.

\bibitem[Zhu et~al.(2024)]{zhu2024scanreason}
Chenming Zhu, Tai Wang, Wenwei Zhang, Kai Chen, and Xihui Liu.
\newblock ScanReason: Empowering 3D visual grounding with reasoning capabilities.
\newblock In \emph{ECCV}, 2024.

\end{thebibliography}
\end{document}